\documentclass[conference,10pt]{IEEEtran}
\usepackage{blkarray}                                      
\usepackage[noend]{algpseudocode}
\usepackage{algorithm}
\usepackage{graphicx}                                      
\usepackage{amsmath}
\usepackage{amssymb}
\usepackage{amsfonts}
\usepackage{amsthm}
\usepackage[mathcal]{eucal}
\usepackage{booktabs}
\usepackage{enumerate}
\usepackage{multirow}
\usepackage[subrefformat=parens,farskip=0pt,justification=centering]{subfig}
\usepackage{color}
\usepackage{cite}                                          
\usepackage{comment}                                       
\usepackage{soul}                                          
\soulregister\cite7
\soulregister\ref7
\soulregister\pageref7
\usepackage{etoolbox}                                      
\usepackage{url}
\usepackage{nth}                                           
\usepackage{bm}                                            
\usepackage{courier}
\usepackage{balance}
\usepackage{threeparttable}
\usepackage[bookmarks=false]{hyperref}
\hypersetup{
    colorlinks = true,
    citecolor  = blue,
    linkcolor  = blue,
    urlcolor   = blue,
}
\usepackage{tikz}
\usetikzlibrary{positioning,calc,fit,decorations.pathmorphing}
\usepackage{filecontents}                                  
\usepackage{pgfplots}
\usepackage{pgfplotstable}
\pgfplotsset{compat=newest}
\usepackage{caption}
\usepackage{cleveref}
\Crefformat{figure}{Fig.~#2#1#3}                           
\Crefname{subfigure}{Fig.}{Figs.}
\Crefname{figure}{Fig.}{Figs.}
\Crefformat{table}{TABLE~#2#1#3}                           
\definecolor{CUHKorange}{RGB}{244,106,18} 
\definecolor{CUHKblue}{RGB}{0,111,190}    
\definecolor{CUHKgreen}{RGB}{0,127,128}   
\definecolor{CUHKred}{RGB}{228,46,36}     
\definecolor{CUHKyellow}{RGB}{198,148,34} 
\definecolor{CUHKdark}{RGB}{114,44,114}   
\definecolor{CUHKmiddle}{RGB}{144,44,144} 
\definecolor{CUHKlight}{RGB}{167,44,167} 

\renewcommand{\vec}[1]{\boldsymbol{#1}}    

\algrenewcommand\textproc{\texttt}

\makeatletter
\let\OldStatex\Statex
\renewcommand{\Statex}[1][3]{%
  \setlength\@tempdima{\algorithmicindent}%
  \OldStatex\hskip\dimexpr#1\@tempdima\relax
}
\makeatother

\RequirePackage[normalem]{ulem} 
\RequirePackage{color}\definecolor{RED}{rgb}{1,0,0}\definecolor{BLUE}{rgb}{0,0,1} 

\graphicspath{{./figs/}}

\usepackage[export]{adjustbox}
\definecolor{myblue}{RGB}{154, 164, 208}
\definecolor{myred}{RGB}{242, 200, 195}

\newcommand{\ie}{\textit{i}.\textit{e}., }

\newif\ifshowfig
\showfigtrue

\begin{document}
\date{}

\title{
    DevelSet: \underline{D}eep Neural L\underline{evel} \underline{Set} for \\ Instant Mask Optimization
}

\iftrue
\author{
    Guojin Chen, \quad
    Ziyang Yu, \quad
    Hongduo Liu, \quad
    Yuzhe Ma, \quad
    Bei Yu \\
    The Chinese University of Hong Kong \\
}
\fi

\maketitle

\begin{abstract}
With the feature size continuously shrinking in advanced technology nodes,
mask optimization is increasingly crucial in the conventional design flow,
accompanied by an explosive growth in prohibitive computational
overhead in optical proximity correction (OPC) methods.
Recently, inverse lithography technique (ILT) has drawn significant attention
and is becoming prevalent in emerging OPC solutions.
However, ILT methods are either time-consuming or in weak performance of mask printability and manufacturability.
In this paper, we present DevelSet, a GPU and deep neural network (DNN) accelerated level set OPC framework for metal layer.
We first improve the conventional level set-based ILT algorithm by introducing the curvature term to reduce mask complexity and applying GPU acceleration to overcome computational bottlenecks.
To further enhance printability and fast iterative convergence,
we propose a novel deep neural network delicately designed with level set intrinsic principles to facilitate the joint optimization of DNN and GPU accelerated level set optimizer.
Experimental results show that DevelSet framework surpasses the state-of-the-art methods in printability and boost the runtime performance achieving instant level (around 1 second).
\end{abstract}

\section{Introduction}

As minimum feature size continues to shrink,
the optical diffraction and proximity effects in lithography become not negligible,
which could seriously degrade the yield of integrated circuits.
To compensate for pattern distortion and improve process window in the lithography process,
optical proximity correction (OPC) is used to ensure pattern transfer fidelity.
Typical OPC approaches encompass rule-based methods~\cite{OPC-ISQED2000-Park},
model-based methods~\cite{OPC-ICCAD2014-Awad, OPC-TCAD2016-Su},
inverse lithography techniques~\cite{OPC-DAC2014-Gao, OPC-ICCAD2017-Ma},
and DNN-based methods\cite{OPC-TCAD2020-Yang, chen2020damo, NEURAL-ILT-ICCAD2020-Jiang}.

In model-based OPC procedure, the edges of the initial mask are fragmented into segments,
which are moved iteratively under the guidance of lithography simulation.
Inverse lithography techniques (ILT) also leverage rigorous simulation to perform mask printability enhancement.
Moreover, ILT can achieve pixel-level optimization and thus find a better solution
in a larger solution space by seeing mask optimization as an inverse imaging problem.
Gao \textit{et al.}~\cite{OPC-DAC2014-Gao} derived a closed-form gradients descent algorithm through direct edge placement error and process window optimization.
In recent years, DNN-based methods have drawn great attention as they can attain significant speedup
while preserve comparable mask printability by incorporating previous experience.
Yang \textit{et al.}~\cite{OPC-TCAD2020-Yang} proposed a generative model to produce an initial mask solution,
which greatly lowers the number of iterations required in traditional ILT methods.
\ifshowfig
\begin{figure}[tb!]
    \vspace{-.3in}
    \centering
    \subfloat[]{ \includegraphics[width=.76\linewidth]{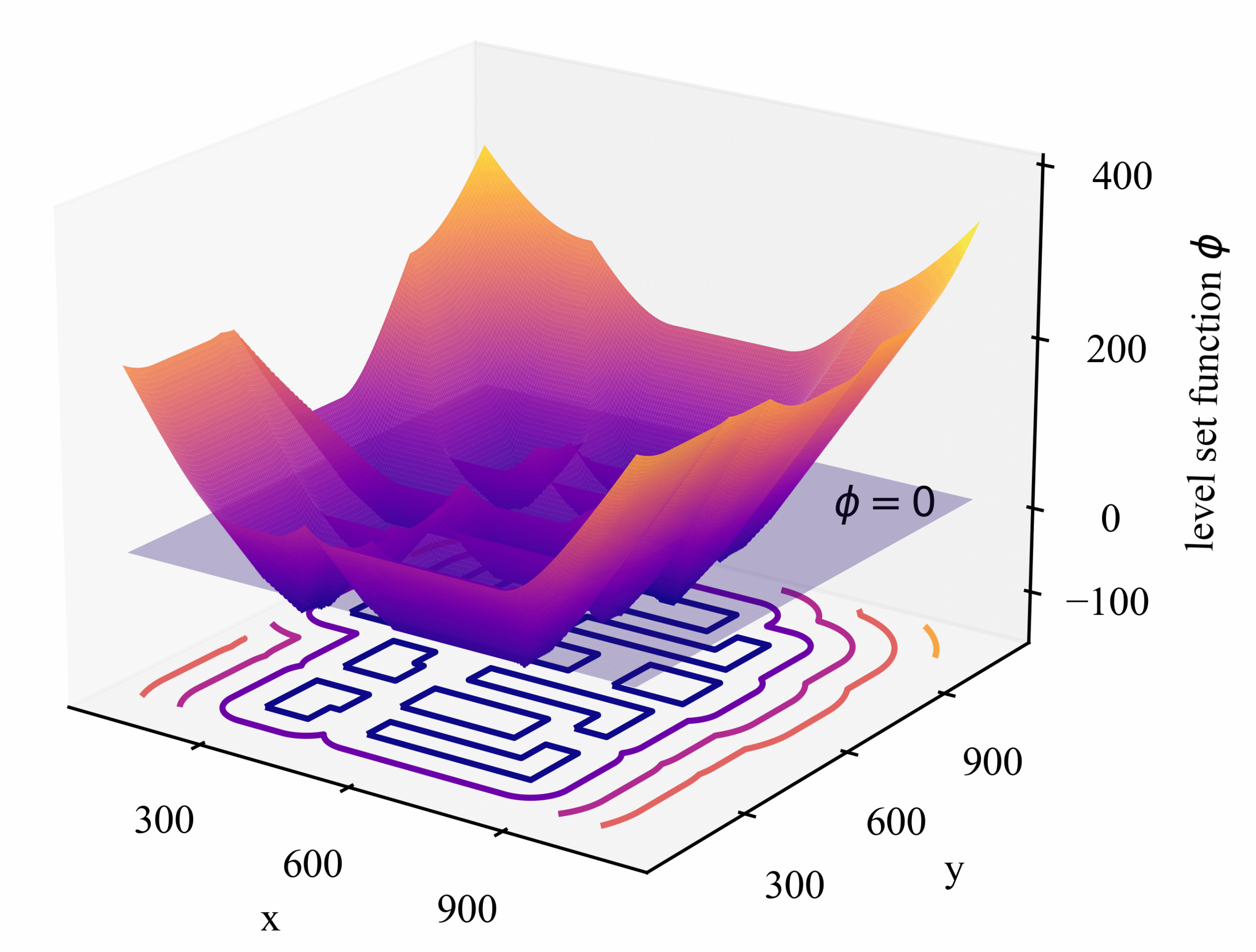} \label{fig:ls_vis_phi}}
    \subfloat[]{ \includegraphics[width=.2\linewidth]{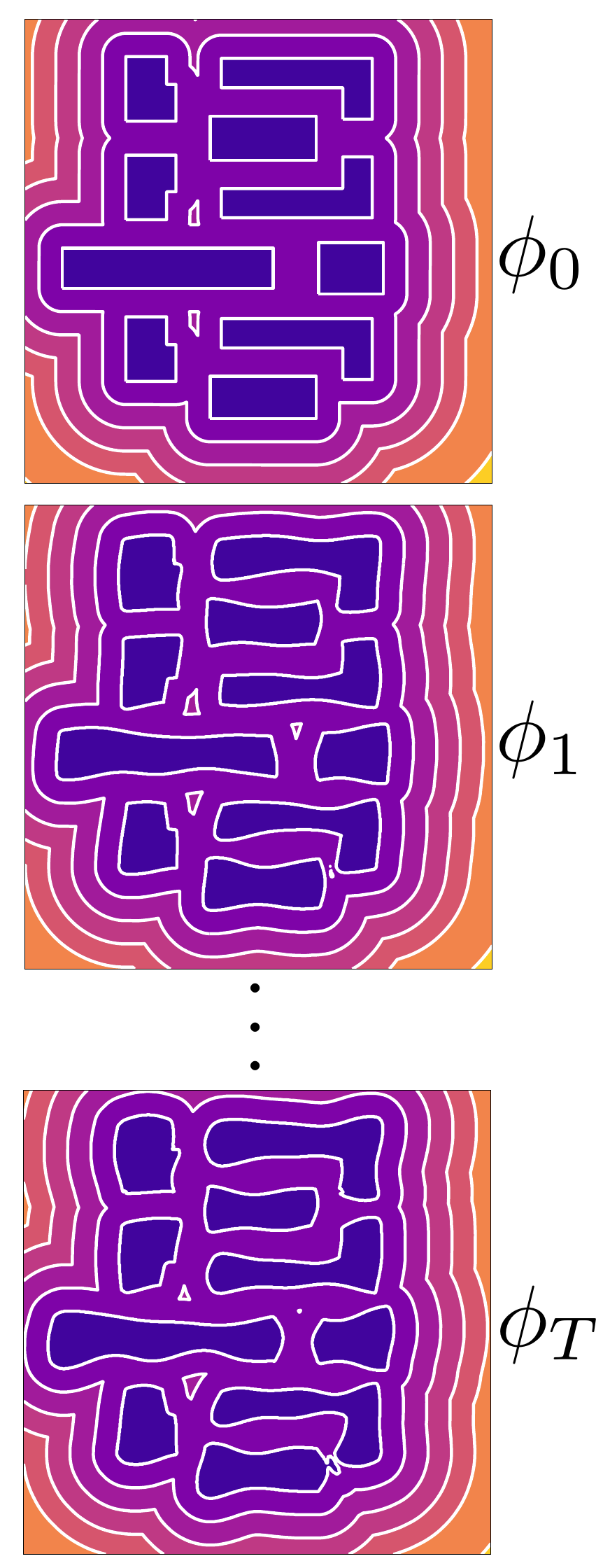} \label{fig:ls_vis_evo}}
    \caption{
      (a) The 3D illustration of level set function $\phi$.
      The mask is shaped as the cross-section of the level set continuum with the zero plane.
      The contours on the x-y plane are the projected level set.
      (b) The level set evolution process.
    }
    \label{fig:ls_vis}
\end{figure}
\fi


In the past decades, level set-based ILT methods have been actively explored as a feasible alternative to pixel-based ILT methods in OPC tasks.
As illustrated in \Cref{fig:ls_vis}, the implicit representation of level set method is naturally more effective in dealing with complex topology changes and lithography development~\cite{sethian1997overview}.
Shen \textit{et al.}~\cite{OPC-OPEX2009-Shen} solved the inverse lithography problem using a level set time-dependent model with finite difference schemes.
Yu \textit{et al.}~\cite{OPC-TR2020-Yu} proposed a momentum-based conjugate gradient (CG) method and accelerated the level set evolution with GPU-enabled Fast Fourier Transform (FFT) algorithm.

Briefly, there are two main approaches for the inverse lithography techniques: parametric and implicit.
The parametric methods~\cite{OPC-DAC2014-Gao, OPC-ICCAD2017-Ma, OPC-TCAD2020-Yang, chen2020damo, NEURAL-ILT-ICCAD2020-Jiang} use pixel-wise tensors to generate the mask (\Cref{fig:test8_pixel}).
While the implicit approaches represent the mask as a zero level set cross-section~\cite{OPC-OPEX2009-Shen, shen2011robust, OPC-JVSTB2013-Lv, geng2015fast, OPC-TR2020-Yu} (\Cref{fig:test8_ls}).
So far, due to the simplicity and flexibility of pixel-based gradient descent methods, the parametric methods have been thoroughly researched through the perspectives of objective function, optimization method, and the DNN acceleration, achieving state-of-the-art (SOTA) runtime performance and mask print fidelity.
However, as depicted in \Cref{fig:test8_mask_pixel}, the parametric methods inevitably generate unnecessary isolated stains or edge glitches with zigzagging and tortuous complex mask boundaries
while the level set implicit representation is accomplished in mask boundary continuity and curvature control (\Cref{fig:test8_mask_ls}).
Unfortunately, due to the extra computational overhead introduced by the level set evolution, the application of level set-based ILT method has been greatly underestimated.

With the rapid development of GPU and deep learning, the progressive potential of level set-based ILT methods should be reconsidered.
Motivated by these issues, we present the DevelSet framework, which contains two parts.
The GPU accelerated level set optimizer (DevelSet-Optimizer) and the deep level set neural network (DevelSet-Net).
Following the improvements proposed by the previous work, DevelSet-Optimizer (DSO) incorporates the curvature term into level set-based ILT to reduce mask complexity
and develops a set of GPU friendly algorithm to overcome the computational overhead.
DevelSet-Net (DSN) is designed to provide better initial solutions by leveraging the fast inference ability of neural network and compensate DSO for the curvature cost by applying a novel modulation branch.
The DevelSet framework benefits from end-to-end joint optimization of DSN and DSO, achieving SOTA fast convergence and mask printability.
Our main contributions are:
\begin{itemize}
  \item We propose DevelSet, an improved level set-based ILT framework with CUDA and DNN acceleration.
  \item We firstly introduce curvature term into level set-based ILT methods to reduce mask complexity and leverage GPU to perform all the calculations.
  \item We are the first to integrate level set into deep neural network for an end-to-end joint mask optimization flow.
  \item We design a novel multi-branch neural network architecture with level set embeddings to further boost the performance and improve mask printability.
  \item Experimental results show that DevelSet achieves SOTA mask printability with predominant runtime advantage for instant mask optimization \ie around 1 second.
\end{itemize}

The rest of the paper is organized as follows: \Cref{sec:prelim} lists some preliminaries about level set algorithms and mask optimization methods.
\Cref{sec:algo} details the DevelSet algorithm. \Cref{sec:exp_results} presents our experimental results, followed by a conclusion in \Cref{sec:conclu}.

\ifshowfig
\begin{figure}[tb!]
    \centering
    \subfloat[]{ \includegraphics[height=.416\linewidth]{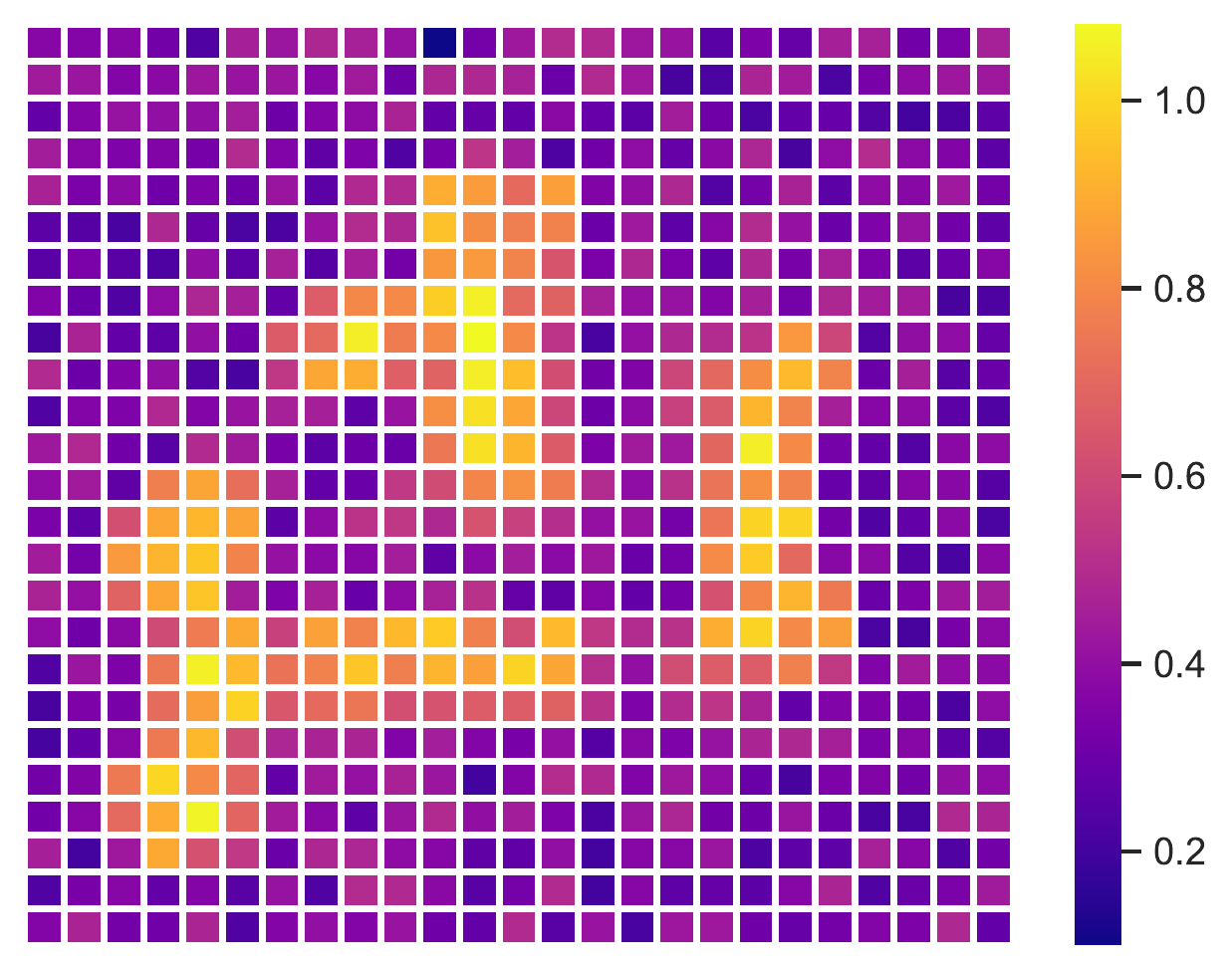} \label{fig:test8_pixel}}
    \subfloat[]{ \includegraphics[height=.416\linewidth]{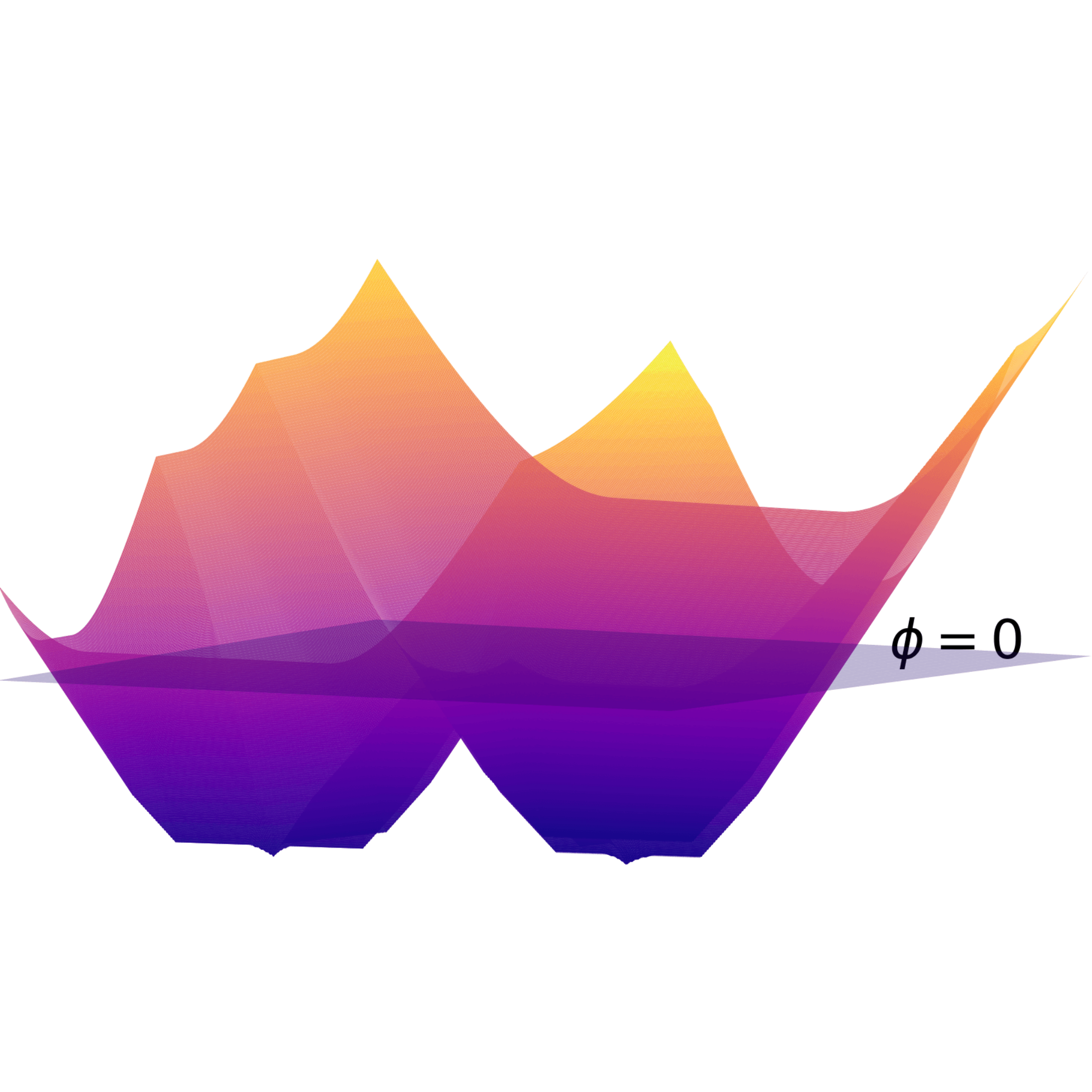} \label{fig:test8_ls}} \\
    \subfloat[]{ \includegraphics[height=.426\linewidth]{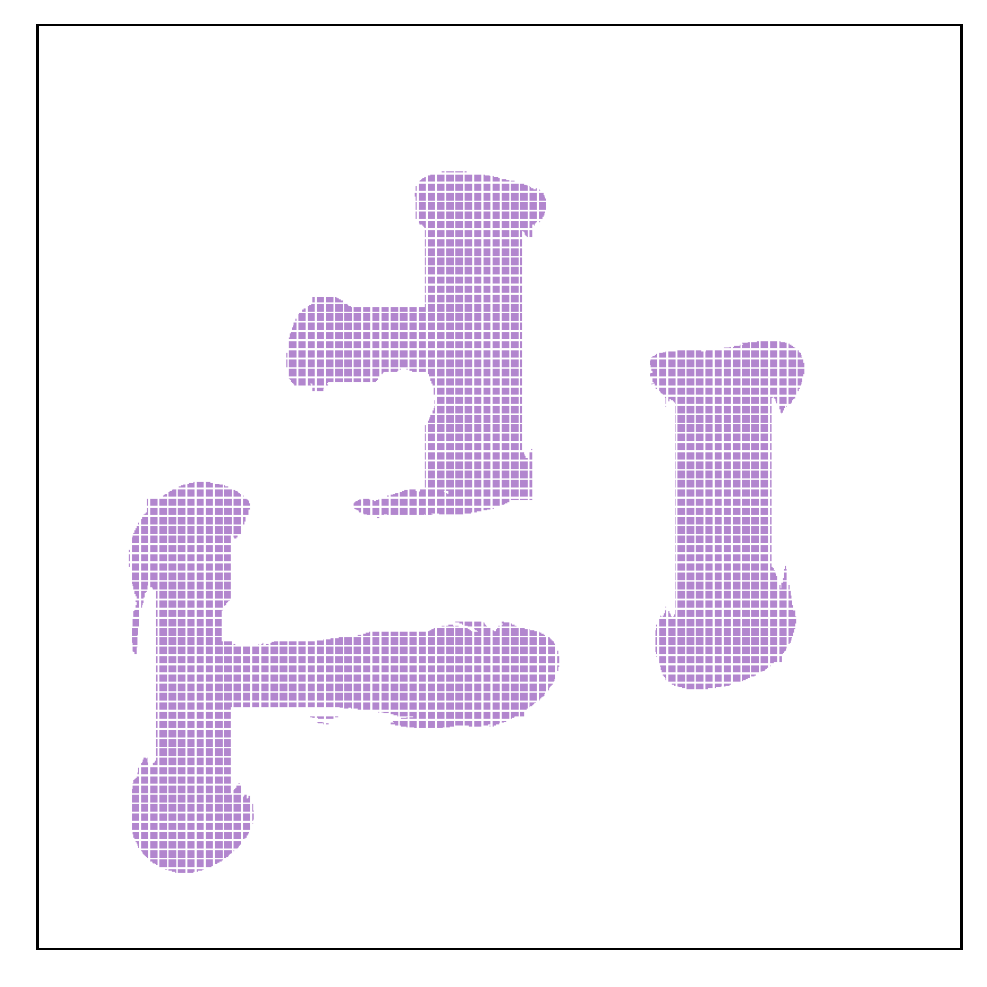} \label{fig:test8_mask_pixel}}
    \subfloat[]{ \includegraphics[height=.426\linewidth]{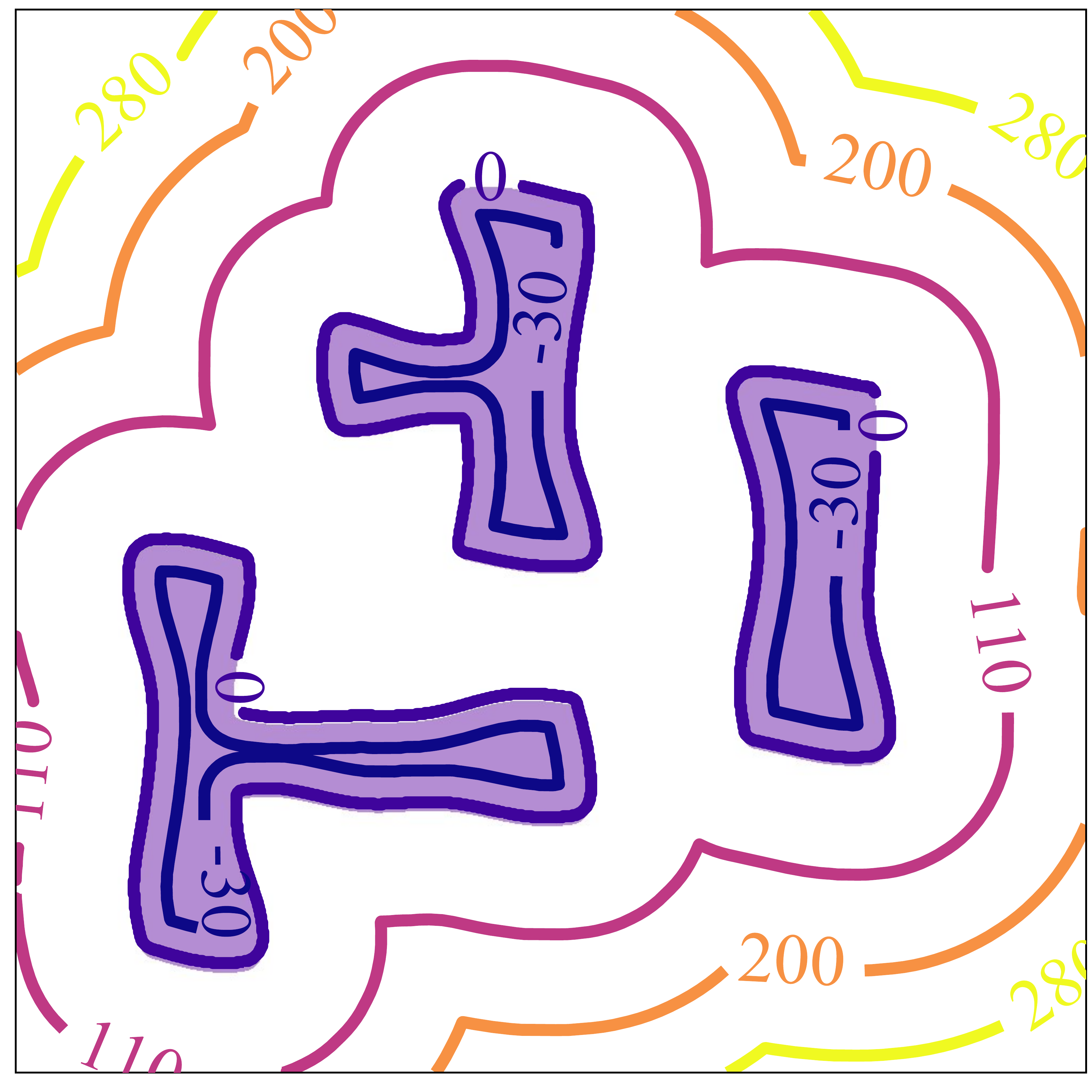} \label{fig:test8_mask_ls}}
    \caption{
        Comparison of pixel-based ILT and level set-based ILT.
        (a) Intensity matrix of pixel-based ILT;
        (b) Level set-based ILT;
        (c) Mask generated by pixel-wise intensity threshold;
        (d) Mask generated by zero level set.
    }
    \label{fig:pixel_vs_ls}
\end{figure}
\fi


\section{Preliminaries}
\label{sec:prelim}
In this section, we will introduce some concepts and background related to this work.
Following the traditions, we denote $\mathbf{Z}_\mathrm{t}$, $\mathbf{M}$, and $\mathbf{I}$, as the target layout, mask image, and intensity of aerial image respectively.
$\mathbf{Z}$, $\mathbf{Z}_{in}$, and $\mathbf{Z}_{out}$ are the wafer image under the nominal/min/max process conditions.
We use $\mathbf{H}$ for lithography kernels and $\phi$ for the level set function (LSF).

\subsection{Level Set-Based ILT Algorithms}

Level set as a mathematical technique is pioneered by Osher \textit{et al.}~\cite{osher1988fronts}.
Recently, it has been actively explored as a feasible alternative to tackle the ILT problem.
Let $\mathcal{C} : \Omega \rightarrow \mathbb{R}^2$ denote a parametric curve in $2$D space $\Omega$,
the level set method implicitly represents the boundary of the mask using zero crossing of a LSF $\phi(x,y): \Omega \rightarrow \mathbb{R}$:
\begin{equation}
  \mathcal{C}=\{(x, y) \mid \phi(x, y)=0\} .
\end{equation}
As depicted in \Cref{fig:ls_vis_evo}, the mask optimization process can be viewed as the evolution along the descent of the LSF $\phi$.
The commonly used LSF $\phi$ is the signed distance function (SDF),
\begin{equation}
  \label{eq:sdf}
  \phi_{\mathrm{SDF}}(x, y)=\left\{\begin{array}{ll}
  -d(x, y), & \mathrm{if} (x, y) \in \mathrm{inside}(\mathcal{C}), \\
  0, & \mathrm{if} (x, y) \in \mathcal{C}, \\
  d(x, y), & \mathrm {if} (x, y) \in \mathrm{outside}(\mathcal{C}),
  \end{array}\right.
\end{equation}
where $d(x, y)$ is the minimum Euclidean distance from point $(x, y)$ to the parametric curve $\mathcal{C}$.
As illustrated in \Cref{fig:test8_mask_ls}, the contours are labeled with its SDF values,
and the $\mathcal{C}$ is the contour labeled by 0.
Now the mask image $\mathbf{M}$ can be represented by $\phi$ as
\begin{equation}
  \label{eq:phi2mask}
  \mathbf{M}(x, y)=\left\{\begin{array}{ll}
  1, & \mathrm{ if }\ \phi(x, y) \leq 0 \\
  0, & \mathrm{ if }\ \phi(x, y) > 0 .
  \end{array}\right.
\end{equation}
During the evolution, the mask boundary $\mathcal{C}(t)$ changes over time $t \in \mathbb{R}$, the curve evolution then can be formally defined as
\begin{equation}
  \label{eq:partial_c_t}
  \frac{\partial \mathcal{C}(t)}{\partial t}=v \vec{n},
\end{equation}
where $\vec{n} = \frac{\nabla \phi}{\vert\nabla\phi\vert}$ is the unit vector in the outward normal direction of the curve $\mathcal{C}$
and $v$ indicates the velocity along the normal direction.
We use the zero level set to implicitly represent the mask boundary, thus: $\phi(\mathcal{C}(t),t) = 0$.
The chain rule gives us,
\begin{equation}
  \label{eq:partial_chain_rule}
  \begin{aligned}
    \frac{\partial \phi(\mathcal{C}(t), t)}{\partial t} =0 \rightarrow
    \frac{\partial \phi}{\partial \mathcal{C}(t)} \frac{\partial \mathcal{C}(t)}{\partial t}+\frac{\partial \phi }{\partial t} = 0 .
  \end{aligned}
\end{equation}
Consider all the points on the evolving front $\mathcal{C}(t)$, $\frac{\partial \phi}{\partial \mathcal{C}} = \nabla\phi$, combining the \Cref{eq:partial_c_t} and \Cref{eq:partial_chain_rule},
the motion equation of LSF $\frac{\partial\phi}{\partial t}$ can be formally expressed by
\begin{equation}
  \label{eq:phi_v}
  \frac{\partial \phi}{\partial t}= -v\vert\nabla\phi\vert.
\end{equation}

\Cref{eq:phi_v} is a partial differential equation (PDE), once the level set $\phi$ and velocity $v$ are defined,
the first-order derivative in space and time of \Cref{eq:phi_v} can be approximated using finite difference techniques.
Evolution of LSF $\phi(x, y, t)$ can be performed iteratively.
We use $\phi_i(x, y)$ to denote $\phi(x, y, t_i)$ for simplicity.
For $i \in \{0, 1, 2 \dots T-1\}$, the $i^{th}$-step update is
\begin{equation}
  \label{eq:levelset_evolution}
  \phi_{i+1}(x, y)=\phi_{i}(x, y)+\Delta t \frac{\partial \phi_{i}}{\partial t} ,
\end{equation}
where $\Delta t$ is the time step, $\phi_0(x,y)$ is the initial LSF,
and the $\phi_T(x, y)$ is the corresponding output LSF after $T$ evolution steps.
As shown in \Cref{fig:ls_vis_evo}, we can obtain the optimized mask after $T$ steps by applying the \Cref{eq:phi2mask}.

\subsection{The Lithography Simulation Model}
\label{sec:lithomodel}

During the conventional lithography process, the input mask $\mathbf{M}$ is transformed through an optical projection system into the aerial image. 
The distribution of aerial light intensity $\mathbf{I}$ floating on the wafer forms the printed image $\mathbf{Z}$.
The optical projection system can be expressed mathematically using Hopskin's diffraction model~\cite{OPC-RSL1951-Hopkins}.
The sum of coherent systems (SOCS) can roughly estimate Hopskin's diffraction model by performing singular value decomposition, the optical projection process is then replaced by a set of coherent kernels. 
The intensity of aerial image $\mathbf{I}$ can be represented by convolving the mask $\mathbf{M}$ and a set of optical kernels $\mathbf{H}$,
\begin{equation}
  \mathbf{I}(x, y) = \sum_{i=1}^{N^2} \sigma_i | \mathbf{M}(x, y) \otimes h_i(x, y)|^2 .
\end{equation}
Here $\otimes$ denotes the convolution operation,
and $h_i$ is the $i^{th}$ kernel of the optical kernel set $\mathbf{H}$
and $\sigma_i$ is the corresponding weight of the coherent system.
The $N_k^{th}$ order approximation to the partially coherent system can be obtained by,
\begin{equation}
  \mathbf{I}(x, y) \approx \sum_{i=1}^{N_{k}} \sigma_{i}\left|\mathbf{M}(x, y) \otimes h_{i}(x, y)\right|^{2},
\end{equation}
where $N_k = 24$ in our implementation.
After optical simulation, the aerial image undergoes a resist model
to estimate the final printed  shape on wafer.
For methodology verification and also for simplicity,
we adopt the constant threshold resist model which is consistent with the ICCAD 2013 contest settings~\cite{OPC-ICCAD2013-Banerjee}.
As depicted in \Cref{fig:test8_mask_pixel}, given the print threshold $I_{th}$, the printed wafer image can be expressed as:
  \begin{align}
    \label{eq:binary_intensity}
    \begin{split}
      \mathbf{Z}= \left \{
        \begin{array}{lr}
          1,  & \mathrm{if} \ \mathbf{I} \ge I_{th}, \\
          0,  & \mathrm{if} \ \mathbf{I} < I_{th}.
        \end{array}
        \right.
      \end{split}
    \end{align}


\subsection{Mask Printability and Mask Manufacturability}
Mask printability represents the quality of the printed patterns generated from the optimized mask.
In this paper, we use squared $L_2$ error and process variation band (PVBand) as two typical metrics to evaluate mask printability.
Moreover, the mask fracturing shot count proposed in Neural-ILT~\cite{NEURAL-ILT-ICCAD2020-Jiang} is also applied in this work to evaluate mask complexity and manufacturability.

\subsubsection{Squared $L_2$ error} Given the wafer image $\mathbf{Z}$ and target image $\mathbf{Z}_\mathrm{t}$, the squared $L_2$ error is calculated by:
$\left\|\mathbf{Z}-\mathbf{Z}_{\mathrm{t}}\right\|_{2}^{2}$.

\subsubsection{PVBand} Process variation band (PVBand) is the bitwise-XOR region among all the printed patterns under different process conditions.
In our work, for simplicity, we calculate the PVBand under two extreme conditions, one at nominal condition with $+2\%$ dose and the other one at defocus and $-2\%$ dose.
A mask is more robust if its PVBand area is smaller.

\subsubsection{Mask Fracturing Shot Count} Many conventional pixel-based ILT methods tend to optimize the mask only to improve mask printability.
However, most of these optimized masks contain plenty of tiny irregular sub-features, which increase the difficulty for mask manufacture.
In this work, we use shot count to evaluate the mask manufacturability.
An evaluated mask $\mathbf{M}$ can be fractured into a set of small rectangles which could replicate exactly the original mask.
Mask fracturing shot count stands for the number of fractured rectangles.


\section{Deep Neural Level Set Algorithms}
\label{sec:algo}

As depicted in \Cref{fig:develset_flow}, the proposed DevelSet framework consists of two parts, DevelSet-Optimizer (DSO) and DevelSet-Net (DSN).
In this section, we will first introduce the improved level set-based ILT algorithm of DSO in \Cref{subsec:dso} applying curvature term to improve the mask manufacturability,
along with the full implementation in CUDA platform by combining the mechanism of GPU parallelism with the numerical setting of level set.
Then a novel multi-branch neural network \ie DSN is proposed in \Cref{subsec:dsn} to: \textit{1)} provide a better initial LSF for DSO to reduce the total iterations,
\textit{2)} predict a weighted matrix to selectively regularize the mask boundary and compensate for mask printability loss caused by the curvature term.
Finally, we perform the end-to-end joint optimization for DevelSet in \Cref{subsec:develset} to accomplish instant mask optimization with higher mask printability and lower mask complexity.

\ifshowfig
\begin{figure*}[tb!]
  \centering
  \includegraphics[width=.99\linewidth]{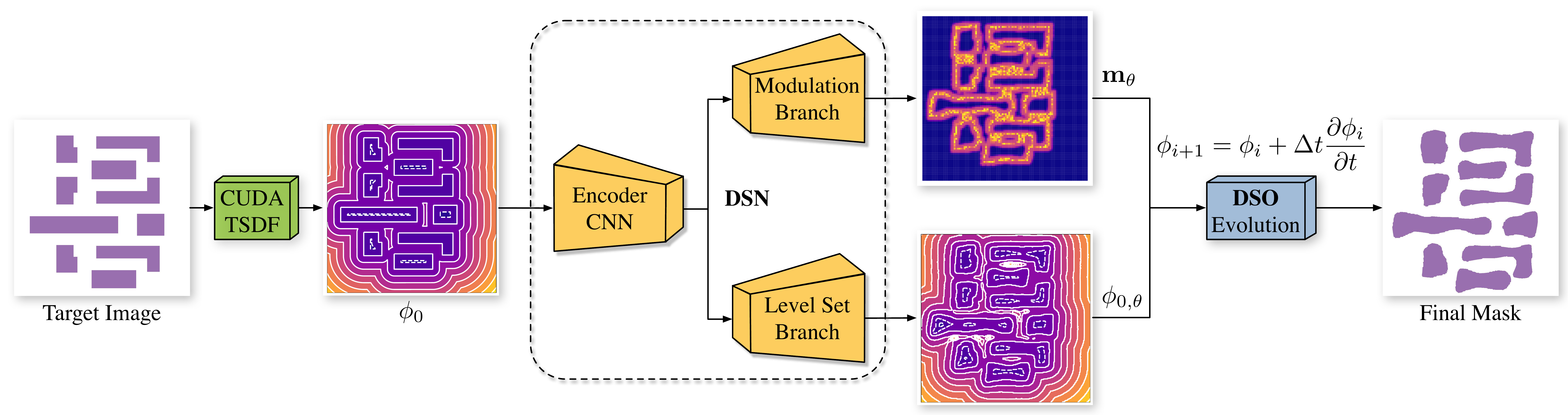}
  \caption{Overview of DevelSet framework with the end-to-end joint optimization flow of DSN and DSO.}
  \label{fig:develset_flow}
\end{figure*}
\fi

\subsection{DevelSet-Optimizer~(DSO)}
\label{subsec:dso}
The objective of the conventional ILT-based OPC method is to find an optimized mask $\mathbf{M}_\mathrm{opt} = \mathcal{L}^{-1}(\mathbf{Z}_\mathrm{t};\mathbf{P}_\mathrm{nom})$,
where $\mathbf{Z}_\mathrm{t}$ is the design target and the $\mathcal{L}\left(\cdot ; \mathbf{P}_{\mathrm {nom }}\right)$ denotes the forward lithography process under the nominal condition.
The pixel-based ILT methods represent the intensity wafer image as the pixel-wise parameters and then update the mask pixel-by-pixel with the guidance of the inverse gradient from the lithography model.
While the level set-based ILT methods regard the optimization process as the evolution of the level set continuum, the mask is formulated by the cross-section of zero height plane and the level set continuum.
Mathematically, the level set continuum is represented with LSF $\phi$, the evolution procedure can be expressed by \Cref{eq:levelset_evolution}.
The SOTA pixel-based method Neural-ILT~\cite{NEURAL-ILT-ICCAD2020-Jiang} brings the ILT to the on-neural-network training solution with a CUDA accelerated lithography simulator, achieving a breakthrough in runtime boosting.
It applies a mask complexity refinement layer and domain knowledge re-training to eliminate the complex shape in masks thus reducing mask complexity.
Nonetheless, Neural-ILT sacrifices mask printability that the sum of $L_2$ and PVBand is worse than the previous learning-based work PGAN-OPC~\cite{OPC-TCAD2020-Yang}.
The SOTA level set-based method GLS-ILT~\cite{OPC-TR2020-Yu} leveraged GPU accelerated FFT to speed up the momentum-based algorithm.
However, there is still room for acceleration since GLS-ILT only accelerates the FFT module and much runtime is wasted on CPU and GPU transfers.

DSO is a CUDA accelerated iterative mask optimizer applying our improved level set algorithm.
By controlling the curvature term and transferring all the computations to the GPU, DSO obtains mask printability superiority and breakthrough in runtime performance simultaneously.
As revealed in \Cref{eq:levelset_evolution}, the key to leverage level set methods in mask optimization tasks is the definition of the LSF $\phi$ and the velocity term $v$.
In DSO, we carefully design the LSF $\phi$, and then integrate the ILT-based gradient as the velocity term $v$.
Moreover, we introduce the curvature term to improve mask printability and reduce mask complexity.


\subsubsection{The Improved Level Set-Based ILT}

\textbf{Truncated signed distance function}.
The theoretical framework of the level set method is independent of which particular LSF $\phi$ is used.
A popular type of LSF is the signed distance function (SDF) in~\Cref{eq:sdf}.
SDF is Lipschitz continuous and from Rademacher's theorem, it is almost everywhere differentiable.
However, instead of using SDF, we adopt truncated signed distance function (TSDF)
as our LSF with an upper bound $D_u$ and a lower bound $D_l$,
\begin{equation}
    \label{eq:tsdf}
    \phi_{\mathrm{TSDF}} =\left\{\begin{array}{ll}
        D_{u}, & \mathrm{if} \ \phi_{\mathrm{SDF}} > D_u, \\
        \phi_{\mathrm{SDF}}, & \mathrm{if} \  D_{l} \leq \phi_{\mathrm{SDF}} \leq D_u, \\
        D_l , & \mathrm {if} \ \phi_{\mathrm{SDF}} < D_l.
    \end{array}\right.
\end{equation}
In our work, we set $D_u$ to $900$ and $D_l$ to $-100$ according to the design rules of the benchmark.
The TSDF improves the stability of optimization procedure by reducing the variance of the dataset.
More importantly, the TSDF ensures the fast convergence of the DevelSet-Net to enable the end-to-end joint optimization of the whole mask optimization framework.

\textbf{Motion term}.
Motion term  $\frac{\partial\phi}{\partial t}$ is the core component in level set methods determining the evolution process.
According to \Cref{eq:phi_v}, it is dominated by the velocity term $v$.
In our work, the velocity is the gradient back-propagated from the forward lithography simulation process of the partial coherent imaging system.
All the techniques used in the previous pixel-based framework can be seamlessly migrated to the DevelSet framework,
for DSO provides a proxy that allows us to perfectly control the gradient near the mask boundary.
The objective function of DSO consists of the commonly used ILT loss and the PVBand loss,
\begin{equation}
  L_{\mathrm{DSO}} =  \alpha L_{\mathrm{ilt}} + \beta L_{\mathrm{pvb}}.
\end{equation}
\par
To minimize the image difference of target image and nominal image, the ILT loss is given by
\begin{equation}
  L_{\mathrm{ilt}}=\sum_{x=1}^{N} \sum_{y=1}^{N}\left(\mathbf{Z}(x, y)-\mathbf{Z}_{\mathrm{t}}(x, y)\right)^{2},
\end{equation}
where the $\mathbf{Z}_t$ is the target image; $\mathbf{Z}$ is the wafer image after the lithography under the nominal condition; N is the width of the target image.
To enable the evolution process differentiable, the step function in \Cref{eq:binary_intensity} is approximated as
\begin{equation}
  \mathbf{Z} = \frac{1}{1+\exp \left(-\sigma_{z} \times\left(\mathbf{I}-I_{t h}\right)\right)},
\end{equation}
where $\sigma_z$ is the steepness of the sigmoid function. Then the gradient of ILT loss can be expressed as
\begin{equation}
  \label{eq:ilt_gradient}
  \begin{aligned}
  &\frac{\partial L_{\mathrm{ilt}}}{\partial \mathbf{M}}= 2 \times\left(\mathbf{Z}-\mathbf{Z}_{\mathrm{t}}\right) \odot \frac{\partial \mathbf{Z}}{\partial \mathbf{M}} \\
  &=2\sigma_{z} \times\left\{\mathbf{H}^\mathrm{\prime} \otimes\left[\left(\mathbf{Z}-\mathbf{Z}_{\mathrm{t}}\right) \odot \mathbf{Z} \odot(1-\mathbf{Z}) \odot\left(\mathbf{M} \otimes \mathbf{H}^{*}\right)\right]\right.\\
  &\left.+(\mathbf{H}^{\mathrm{\prime}})^{*} \otimes\left[\left(\mathbf{Z}-\mathbf{Z}_{\mathrm{t}}\right) \odot \mathbf{Z} \odot(1-\mathbf{Z}) \odot(\mathbf{M} \otimes \mathbf{H})\right]\right\} ,
  \end{aligned}
\end{equation}
where the $\mathbf{H}^{\prime}$ is the $180^{\circ}$ rotation of the optical kernel set $\mathbf{H}$, and the $\mathbf{H}^{*}$ is the conjugate of $\mathbf{H}$.

To minimize the area of PVBand, we expect the innermost/outermost wafer under min/max process conditions as close to the target image as possible.
The PVBand loss is given by,
\begin{equation}
  L_{\mathrm{pvb}}  = \left(\mathbf{Z}_\mathrm{in}-\mathbf{Z}_{\mathrm{t}}\right)^{2} + \left(\mathbf{Z}_\mathrm{out}-\mathbf{Z}_{\mathrm{t}}\right)^{2} .\\
\end{equation}
The gradient of PVBand loss can be represented as
\begin{equation}
  \label{eq:pvb_gradient}
  \begin{aligned}
    \frac{\partial L_{\mathrm{pvb}}}{\partial \mathbf{M}} =& 2 \times\left(\mathbf{Z}_{\mathrm{in}}-\mathbf{Z}_{\mathrm{t}}\right)\odot\frac{\partial \mathbf{Z}_\mathrm{in}}{\partial \mathbf{M}} \\
                                                          +& 2 \times\left(\mathbf{Z}_{\mathrm{out}}-\mathbf{Z}_{\mathrm{t}}\right)\odot\frac{\partial \mathbf{Z}_\mathrm{out}}{\partial \mathbf{M}}.
  \end{aligned}
\end{equation}
The detailed derivation of \Cref{eq:pvb_gradient} is similar to \Cref{eq:ilt_gradient}.
Now the velocity $v$ is
\begin{equation}
  \label{eq:vel}
  v = \alpha \frac{\partial L_{\mathrm{ilt}}}{\partial \mathbf{M}} + \beta \frac{\partial L_{\mathrm{pvb}}}{\partial \mathbf{M}} .
\end{equation}
And the motion equation is finally derived as
\begin{equation}
  \label{eq:phi_v_grad}
  \frac{\partial \phi_{i}}{\partial t}= -(\alpha \frac{\partial L_{\mathrm{ilt}}}{\partial \mathbf{M}} + \beta \frac{\partial L_{\mathrm{pvb}}}{\partial \mathbf{M}})|\nabla\phi_i|.
\end{equation}

\textbf{Curvature term}.
As revealed in \Cref{eq:levelset_evolution}, the evolution manner of level set is defined by several updating terms,
which can be roughly divided into two categories :
(1) external terms that attract the curve to the desired location-based on the data evidence, such as the inverse lithography gradient or the optimization methods,
and (2) internal regularization terms on the curve shape, \textit{e.g.} curvature and the length of the curvature.
Previous level set-based methods focus on the improvements of the external terms since the internal term such as curvature requires extensive calculations to get the second-order derivatives.
However, with GPU acceleration, DSO takes the maximum advantage of the effective feature in the implicit representation to obtain the curvature of the boundaries,
which is practical to control the smoothness of the front and eliminate the noise points on the mask pattern.
The curvature term is formally defined as
\begin{equation}
  \label{eq:curvature_term}
  \kappa = \lambda \mathbf{m}_\theta \left|\nabla \phi_i\right| \operatorname{div}\left(\frac{\nabla \phi_i}{\left| \nabla \phi_i \right|}\right),
\end{equation}
where $\lambda$ is the curvature weight. However, in OPC tasks,
there may exist sharp corners in some parts of the masks.
Directly apply curvature term on the level set evolution process may harm the lithography results.
Thus, we add a weighted matrix $\mathbf{m}_\theta$ to control the curvature term,
and the subscript $\theta$ denotes the $\mathbf{m}_\theta$ is predicted by the modulation branch parameters of DSN.
We will introduce the modulation branch in \Cref{subsubsec:dsn_arch} detailedly.
Then the level set evolution of DSO can be described as the sum of the motion term and curvature term,
\begin{equation}
  \label{eq:dso_evolution}
  \begin{aligned}
  \frac{\partial \phi_i}{\partial t} =& -(\alpha \frac{\partial L_{\mathrm{ilt}}}{\partial \mathbf{M}} + \beta \frac{\partial L_{\mathrm{pvb}}}{\partial \mathbf{M}})|\nabla\phi_i| \\
  &+  \lambda \mathbf{m}_\theta \left|\nabla \phi_i\right| \operatorname{div}\left(\frac{\nabla \phi_i}{\left| \nabla \phi_i \right|}\right) .
  \end{aligned}
\end{equation}

\subsubsection{The CUDA Implementation of DSO}

Conventional ILT-based mask optimization methods suffer from severe computational overhead,
and the situation grows worse in level set-based methods.
When the new terms are leveraged to improve mask printability,
the new computational cost is also introduced to the already burdensome computation system.
Hence, the major challenge for DSO is to overcome the drawback of high computational effort.
By implementing the entire DSO framework on the CUDA platform,
we find a way to balance efficiency and performance.
Next, we will detail the CUDA implementation of each term in level set algorithm,
as well as engineering tricks to make our DSO framework significantly faster.

\textbf{Numerical settings}.
The level set-based mask optimization methods focus on 2D situation with an image as the input.
The space is discretized by a Cartesian grid with steps $\Delta x, \Delta y$, where the coordinates $(x, y)$ represent the $x^{th}, y^{th}$ pixel in the image.
The first-order derivatives in space and time of \Cref{eq:dso_evolution} can be approximated using finite difference techniques.
We apply weighted essential nonoscillatory (WENO)~\cite{HARTEN1987231} numerical polynomial interpolation method that uses the smoothest possible polynomial interpolation to find $\phi$.
And the first-order and second-order spatial derivatives of $\phi$ can be represented with central differences as
\begin{equation}
  \label{eq:numerical-settings}
  \begin{aligned}
  \nabla \phi_{x} =& \frac{1}{2}(\phi(x+1,\ y)-\phi(x-1,\ y)), \\
  \nabla \phi_{y} =& \frac{1}{2}(\phi(x,\ y+1)-\phi(x,\ y-1)), \\
  \nabla \phi_{xx} =& \phi(x+1,\ y) + \phi(x-1,\ y) - 2 \times \phi(x,\ y), \\
  \nabla \phi_{yy} =& \phi(x,\ y+1) + \phi(x,\ y-1) - 2 \times \phi(x,\ y), \\
  \nabla \phi_{xy} =& \frac{1}{4}\ [(\phi(x+1,\ y+1)-\phi(x-1,\ y+1)) \\
  &-(\phi(x+1,\ y-1)-\phi(x-1,\ y-1))],
  \end{aligned}
\end{equation}
and the curvature term is then computed numerically with
\begin{equation}
  \label{eq:curva_numer}
  \begin{aligned}
    \kappa &= \lambda \mathbf{m}_\theta \left|\nabla \phi_i\right| \operatorname{div}\left(\frac{\nabla \phi_i}{\left| \nabla \phi_i \right|}\right) \\
          &= \lambda \mathbf{m}_\theta \frac{\nabla\phi_{x x} {\nabla\phi_{y}}^{2}-2 \nabla\phi_{y} \nabla\phi_{x} \nabla\phi_{x y}+ \nabla\phi_{y y} {\nabla\phi_{x}}^{2}}{{\nabla\phi_{x}}^{2}+ {\nabla\phi_{y}}^{2}} .
  \end{aligned}
\end{equation}

\textbf{CUDA-based TSDF}.
The first tremendous challenge is to calculate the truncated signed distance function (TSDF) on a given target image (2048 $\times$ 2048), in an extremely short period.
The most celebrated method to calculate signed distance function is the Fast Marching Method introduced by \cite{sethian1996fast}.
Instead of using Fast Marching Method, we have specially designed the TSDF algorithm based on the characteristics of CUDA parallelism.
In DSO, we use the target pattern as the initial mask.
The first step focuses on extracting the boundary segments and calculating the distance towards the boundary using the \texttt{CUDA\_TSDF} function in \Cref{alg:parallel-ls}.
We apply pixel-wise \texttt{Shift} and \texttt{XOR} operation to obtain the mask boundary lines $b_h$, $b_v$ (line 2-5).
Then for all pixels $p$ on mask plates we calculate the distance towards all boundary lines and select the minimum distance for each point in parallel.
Finally, we apply the \Cref{eq:tsdf} to generate the truncated signed distance function (line 6-10).
For a complicated mask generated from the neural network, experimental result shows the \texttt{CUDA\_TSDF} can achieve more than 98\% reduction in TSDF calculation time.



\textbf{CUDA-based geometry gradient and curvature term}.
As demonstrated in \Cref{eq:numerical-settings}, the numerical settings are well compatible with CUDA parallelism.
The spatial derivatives of $\phi$ are calculated in function \texttt{CUDA\_geometry\_gradient} of \Cref{alg:parallel-ls}.
And the curvature term can be calculated with the function \texttt{CUDA\_curvature} with GPU acceleration.
All the operations in \Cref{alg:parallel-ls} such as \texttt{shift} and \texttt{XOR} are pixel-wise independent and can be parallelly performed per pixel per thread,
which not only reduces the total runtime of the DSO but also makes it possible to integrate the level set evolution into neural network.


\textbf{CUDA accelerated lithography simulation}.
According to the previous experimental analysis,
lithography simulation is the most time-consuming part of the mask optimization flow,
since it involves plenty of convolution operations between different kernels and the mask images.
Inspired by Neural-ILT~\cite{NEURAL-ILT-ICCAD2020-Jiang}, we implement our CUDA accelerated lithography simulator and integrate the forward and backward functionalities into the popular machine learning framework \texttt{PyTorch}, with some engineering improvements.
First, the optical kernels and corresponding weights are loaded and pinned in GPU memory throughout the optimization process, all the computations are performed on GPU to reduce the data transfer time from CPU to GPU.
Second, the runtime bottleneck of the CUDA lithography simulator lies on the \texttt{CUDA\_FFT} and \texttt{CUDA\_IFFT} operators.
Our improved \texttt{CUDA\_FFT} operator runs faster than the commonly used \texttt{cuFFT} and the \texttt{torch.fft} libraries.


\begin{algorithm}[h]
  \caption{CUDA Level Set Algorithms}
  \label{alg:parallel-ls}
  \begin{algorithmic}[1]
      \Require Target image $\mathbf{Z}_\mathrm{t}$
      \Function{CUDA\_TSDF}{$\mathbf{Z}_\mathrm{t}$}
      \State $\mathbf{Z}_{tu}, \mathbf{Z}_{td} \gets$ Shift $\mathbf{Z}_\mathrm{t}$ upwards, downwards by 1 pixel;
      \State $\mathbf{Z}_{tl}, \ \mathbf{Z}_{tr} \gets$ Shift $\mathbf{Z}_\mathrm{t}$ leftwards, rightwards by 1 pixel;
      \State $b_{h} \gets (\mathbf{Z}_\mathrm{t}$ \texttt{XOR} $\mathbf{Z}_{tu}) + (\mathbf{Z}_\mathrm{t}$ \texttt{XOR} $\mathbf{Z}_{td})$;
      \State $b_{v} \gets (\mathbf{Z}_\mathrm{t}$ \texttt{XOR} $\mathbf{Z}_{tl}\ ) + (\mathbf{Z}_\mathrm{t}$ \texttt{XOR} $\mathbf{Z}_{tr})$;
      \ForAll{pixels on target image $\mathbf{Z}_\mathrm{t}$}
      \State $d_{ij}$ $\gets$ Distance from pixel ${p}_i$ to boundary ${b}_j$;
      \State $d_i$ \ $\gets$ Minimum distance of point ${p}_i$ in all $d_{ij}$;
      \State $\phi_\mathrm{SDF}\ \ $ $\gets$ SDF matrix from all $d_i$;
      \State $\phi_\mathrm{TSDF}$ $\gets$ TSDF matrix using \Cref{eq:tsdf};
      \EndFor
      \State \Return $\phi_\mathrm{TSDF}$;
      \EndFunction
      \Ensure Truncated Signed Distance Function $\phi_\mathrm{TSDF}$;
      \Statex
      \Require TSDF matrix $\phi_\mathrm{TSDF}$;
      \Function{CUDA\_geometry\_gradient}{$\phi$}
      \State $\phi_{u}, \phi_{d} \gets$ Shift $\phi$ upwards, downwards by 1 pixel;
      \State $\phi_{l}, \ \phi_{r} \gets$ Shift $\phi$ leftwards, rightwards by 1 pixel;
      \State $\nabla \phi_x \gets (\phi_{r} - \phi_{l}) / 2$; $\ \nabla \phi_y \gets (\phi_{u} - \phi_{d}) / 2$;
      \State \Return $\nabla \phi_x, \nabla \phi_y$;
      \EndFunction
      \Ensure Geometry gradient $\nabla \phi_x$, $\nabla \phi_y$;
      \Statex
      \Require TSDF $\phi_\mathrm{TSDF}$, geometry gradient $\nabla \phi_x, \nabla \phi_y$;
      \Function{CUDA\_curvature}{$\phi, \nabla \phi_x, \nabla \phi_y$}
      \State $\nabla \phi_{xx} \gets $ \texttt{CUDA\_geometry\_gradient}$(\nabla \phi_x)$;
      \State $\nabla \phi_{yy} \gets $ \texttt{CUDA\_geometry\_gradient}$(\nabla \phi_y)$;
      \State $\phi_{ul}, \phi_{ur}, \phi_{dl}, \phi_{dr} \gets$ Shift $\phi$ to 4 diagonal directions;
      \State $\nabla \phi_{xy} \gets ((\phi_{ur}-\phi_{ul}) - (\phi_{dr} - \phi_{dl}))/4$;
      \State $\kappa \gets $ Curvature term using \Cref{eq:curva_numer};
      \State \Return $\kappa$;
      \EndFunction
      \Ensure Curvature term $\kappa$;
  \end{algorithmic}
\end{algorithm}

\subsection{DevelSet-Net~(DSN)}
\label{subsec:dsn}

Although the CUDA accelerated DSO framework has achieved a remarkable speedup, there is still much room for improvement.
Given the recent advance of deep learning on OPC,
we propose a novel neural network with level set embeddings to improve efficiency and mask printability.

\subsubsection{Network Architecture and Training}
\label{subsubsec:dsn_arch}
As illustrated in \Cref{fig:develset_flow}, the DSN is a mulit-branch neural network adopting the simple UNet~\cite{U-Net} as the backbone.
Our key contribution is the integration of level set embeddings with the conventional OPC networks.


\textbf{Multi-branch pre-training}.
To utilize the advance of the mulit-branch neural networks, two types of losses are optimized simultaneously,

\begin{equation}
  L_{\mathrm{DSN}}(\theta)=L_{0}(\theta)+ L_{m}(\theta) .
\end{equation}

\textbf{Level set branch supervision}.
As illustrated in \Cref{fig:develset_flow}, different from the tipical OPC networks, the level set branch predicts the initial LSF $\phi_{0,\theta}$ for DSO, instead of the pixel-wise mask.
The mean square error is employed as the objective function,
\begin{equation}
  L_{0}(\theta)=\sum_{(x, y)}\left(\phi_{0, \theta}(x, y)-\phi_{\mathrm{gt}}(x, y)\right)^{2},
\end{equation}
where $\phi_{0,\theta}$ is the predicted LSF with network parameters $\theta$.
The $\phi_{gt}$ is the ground truth LSF generate by DSO.

\begin{table*}[htbp]
    \centering
    \caption{Mask Printability, Complexity Comparison with SOTA.}
    \label{tab:develset_results}
    \setlength{\tabcolsep}{2pt}
    \renewcommand{\arraystretch}{1.2}
    \begin{tabular}{cc|ccc|ccc|ccc|ccc|ccc}
        \toprule
        \multirow{2}{*}{Bench} & \multirow{2}{*}{Area($nm^2$)} & \multicolumn{3}{c|}{ ILT \cite{OPC-DAC2014-Gao} } & \multicolumn{3}{c|}{ GLS-ILT \cite{OPC-TR2020-Yu} } & \multicolumn{3}{c|}{ PGAN-OPC \cite{OPC-TCAD2020-Yang} } & \multicolumn{3}{c|}{ Neural-ILT \cite{NEURAL-ILT-ICCAD2020-Jiang} } & \multicolumn{3}{c}{ DevelSet }  \\
        &        & $L_2$ & PVB & \#shots & $L_2$ & PVB & \#shots & $L_2$ & PVB & \#shots & $L_2$ & PVB & \#shots & $L_2$ & PVB & \#shots \\ \midrule
        \texttt{case1} &215344   &49893 &65534 &2478 &46032 &62693 &1476 &52570 &56267 &931 &50795 &63695 &743 &49142 &59607 &969     \\
        \texttt{case2} &169280   &50369 &48230 &704 &36177 &50642 &861 &42253 &50822 &692 &36969 &60232 &571 &34489 &52012 &743     \\
        \texttt{case3} &213504   &81007 &108608 &2319 &71178 &100945 &2811 &83663 &94498 &1048 &94447 &85358 &791 &93498 &76558 &889     \\
        \texttt{case4} &82560    &20044 &28285 &1165 &16345 &29831 &432 &19965 &28957 &386 &17420 &32287 &209 &18682 &29047 &376     \\
        \texttt{case5} &281958   &44656 &58835 &1836 &47103 &56328 &963 &44733 &59328 &950 &42337 &65536 &631 &44256 &58085 &902     \\
        \texttt{case6} &286234   &57375 &48739 &993 &46205 &51033 &942 &46062 &52845 &836 &39601 &59247 &745 &41730 &53410 &774     \\
        \texttt{case7} &229149   &37221 &43490 &577 &28609 &44953 &548 &26438 &47981 &515 &25424 &50109 &354 &25797 &46606 &527     \\
        \texttt{case8} &128544   &19782 &22846 &504 &19477 &22541 &439 &17690 &23564 &286 &15588 &25826 &467 &15460 &24836 &493     \\
        \texttt{case9} &317581   &55399 &66331 &2045 &52613 &62568 &881 &56125 &65417 &1087 &52304 &68650 &653 &50834 &64950 &932     \\
        \texttt{case10} &102400   &24381 &18097 &380 &22415 &18769 &333 &9990 &19893 &338 &10153 &22443 &423 &10140 &21619 &393     \\   \midrule
        \multicolumn{2}{c|}{Average} &44012.7 &50899.5 &1300.1 &38615.4 &50030.3 &968.6 &39948.9 &49957.2 &706.9 &38503.8 &53338.3 &\textbf{ 558.7 } &\textbf{ 38402.8 } &\textbf{ 48673.0 } &699.8     \\
        \multicolumn{2}{c|}{Ratio} &1.146 &1.046 &1.858 &1.006 &1.028 &1.384 &1.040 &1.026 &1.010 &1.003 &1.096 &\textbf{ 0.798 }&\textbf{ 1.000 }&\textbf{ 1.000 }&1.000     \\  \bottomrule
        \multicolumn{4}{l}{\textit{\ \ $^{\dagger}$\footnotesize{$L_{2}$ and PVB unit: $nm^2$}.}}
    \end{tabular}
\end{table*}

\textbf{Modulation branch supervision}.
During the training process, the modulation branch aims to find the best $\mathbf{m}_{\theta}$ in \Cref{eq:dso_evolution} for curvature term evolution in DSO,
which is a boundary-aware model for detecting the curvature-sensitive areas.
The idea is carried out by shifting the ground-truth TSDF $\phi_{\mathrm{gt}}$ with a set of distance $\Delta h$,
\begin{equation}
  \begin{aligned}
    \tilde{\phi}_{m}(x, y) &= \phi_{\mathrm{gt}}(x, y)+\Delta h, \\
    \tilde{m} &= H(\tilde{\phi}_{m}(x, y)) ,
  \end{aligned}
\end{equation}
where $\Delta h$ is uniformly sampled from $[-20, 20]$, $\tilde{m}$ is a set of $\mathbf{m}_\theta$. $H(\phi)$ is the Heaviside function,
\begin{equation}
  \label{eq:heaviside}
  H(z)=\left\{\begin{array}{l}
  1, z \geq 0, \\
  0, z<0.
  \end{array}\right.
\end{equation}
For every target image $Z_\mathrm{t}$, the ground-truth of modulation branch is
\begin{equation}
  m_\mathrm{gt}=\underset{\tilde{m}}{\operatorname{argmin}}\ L_\mathrm{DSO}.
\end{equation}
During the training, the modulation branch learns to simulate an optimized $\mathbf{m}_\theta$.
As suggested in \cite{chan2001atc}, the simple Heaviside function in \Cref{eq:heaviside} acts on zero level set, which may get stuck in the local minima.
To tackle this, we replace it with the Approximated Heaviside Function~(AHF) with a parameter $\varepsilon$,
\begin{equation}
  \begin{aligned}
  H_{\varepsilon}(\phi) &=\frac{1}{2}\left(1+\frac{2}{\pi} \arctan \left(\frac{\phi}{\varepsilon}\right)\right).
  \end{aligned}
\end{equation}
Thus, the objective function is
\begin{equation}
  L_{m}(\theta)=\sum_{(x, y)}\left(H_{\varepsilon}(\phi_{m, \theta}(x, y))-m_{\mathrm{gt}}(x, y)\right)^{2},
\end{equation}
where $H_{\varepsilon}(\phi_{m, \theta})$ is the output of modulation branch.

\subsection{DevelSet~(DSN+DSO) End to End Joint Optimization}
\label{subsec:develset}
As illustrated in \Cref{fig:develset_flow}, we apply the \texttt{CUDA\_TSDF} function to facilitate the fast transform from pixel-wise target image to LSF $\phi_0$.
After pre-training of the two branches of DSN, we fix all the parameters of DSN
then directly feed the output of level set branch $\phi_{0, \theta}$
and modulation branch $\mathbf{m}_{\theta}$ into the evolution process of DSO to generate the final mask.
We choose the conjugate gradient (CG) method~\cite{OPC-JVSTB2013-Lv} for optimization in DSO,
and follow CFL condition~\cite{OPC-JVSTB2013-Lv} to set the time step $\Delta t  = \eta / \mathrm{max}(\lvert v \rvert)$, where $v$ is evolution velocity in \Cref{eq:vel}, and $\eta$ is CFL condition number.

\section{Experimental Results}
\label{sec:exp_results}

The DevelSet framework is developed with the popular deep learning framework \texttt{PyTorch} and CUDA platform.
All the tests are performed on Linux system with 2.2GHz CPU and a single Nvidia Titan Xp GPU.
The lithography engine is from ICCAD 2013 CAD Contest~\cite{OPC-ICCAD2013-Banerjee},
which also provides the ten industrial M1 designs on 32$nm$ design node as evaluation dataset.
The scripts for shot count evaluation are obtained from the authors of Neural-ILT~\cite{NEURAL-ILT-ICCAD2020-Jiang} to guarantee comparable results.
The training set of DevelSet-Net is obtained from the author of GAN-OPC~\cite{OPC-TCAD2020-Yang}.
We pick $\sigma_{z} = 50$, $N_h = 24$, $\alpha = 1$, $\beta = 7.5$, $\lambda = 0.9$, $\varepsilon = 0.03$, and $\eta = 0.85$ for DevelSet optimization.

\subsection{Comparison with State-of-the-art.}
\subsubsection{Mask Printability and Complexity}

\begin{table}[tb!]
  \setlength{\tabcolsep}{.1pt}
  \begin{tabular}{lcccc}
      \small
    & \texttt{case1} & \texttt{case3} & \texttt{case5} &  \texttt{case9} \\
    (a) PGAN~\cite{OPC-TCAD2020-Yang}         & \includegraphics[width=.21\linewidth,valign=m]{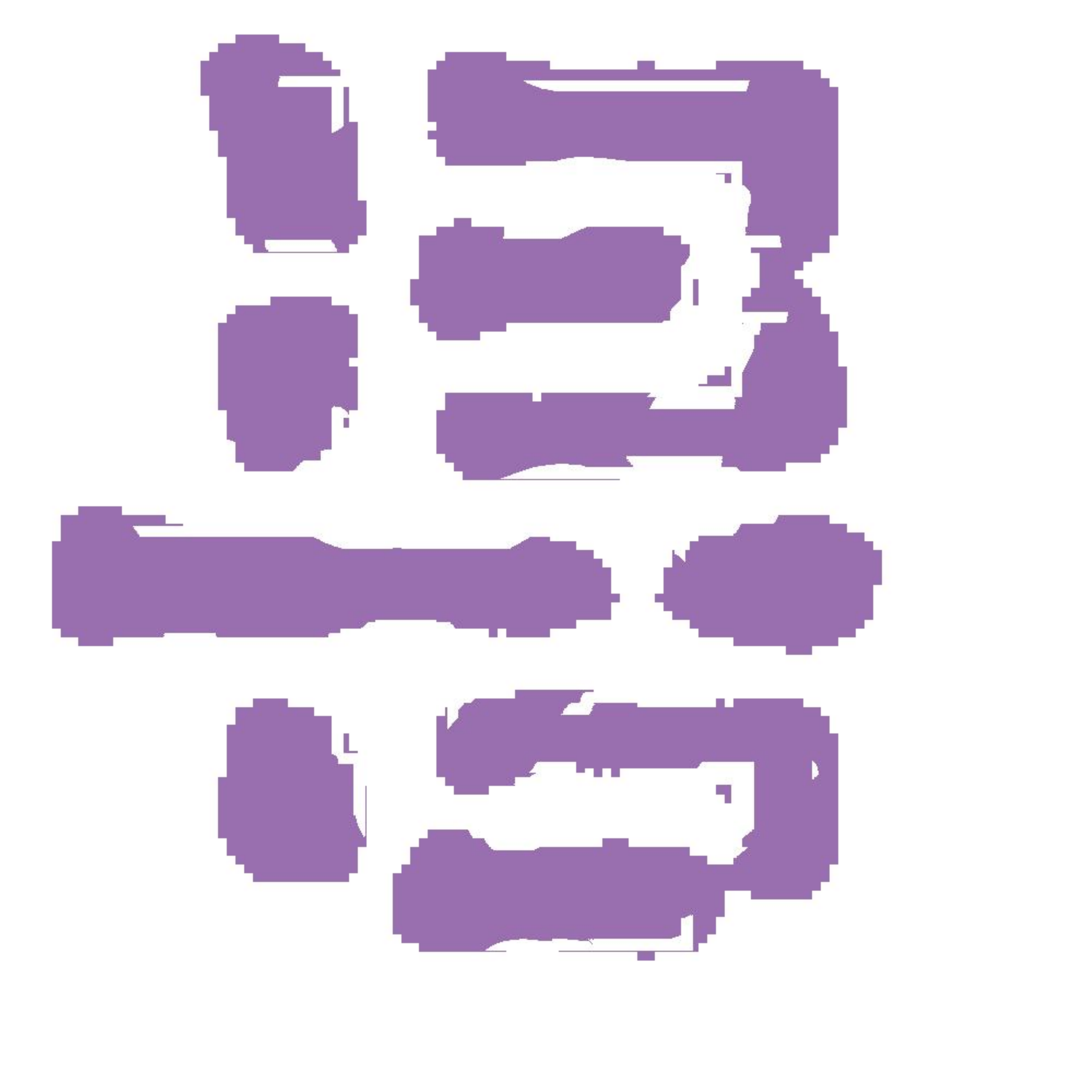} & \includegraphics[width=.21\linewidth,valign=m]{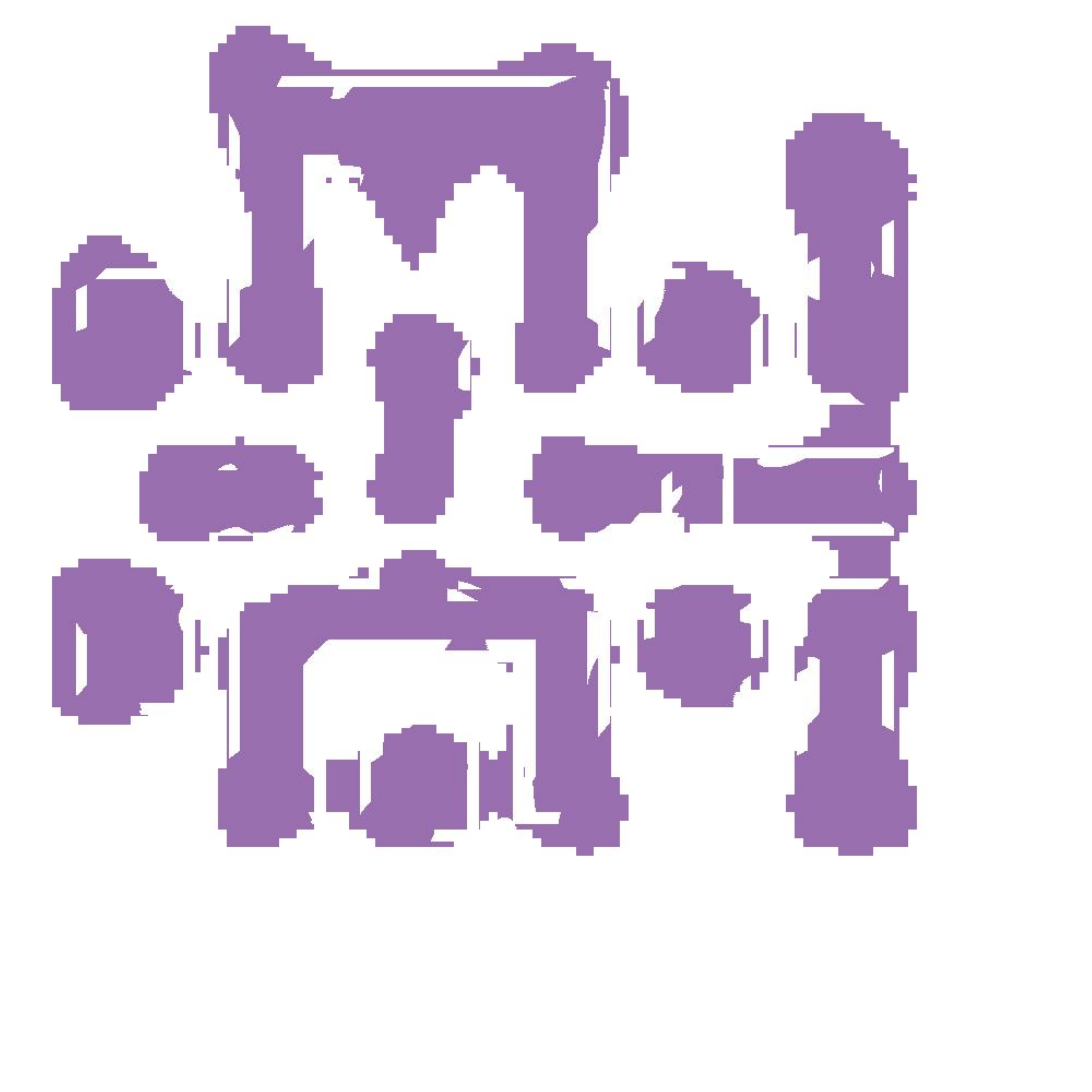} & \includegraphics[width=.21\linewidth,valign=m]{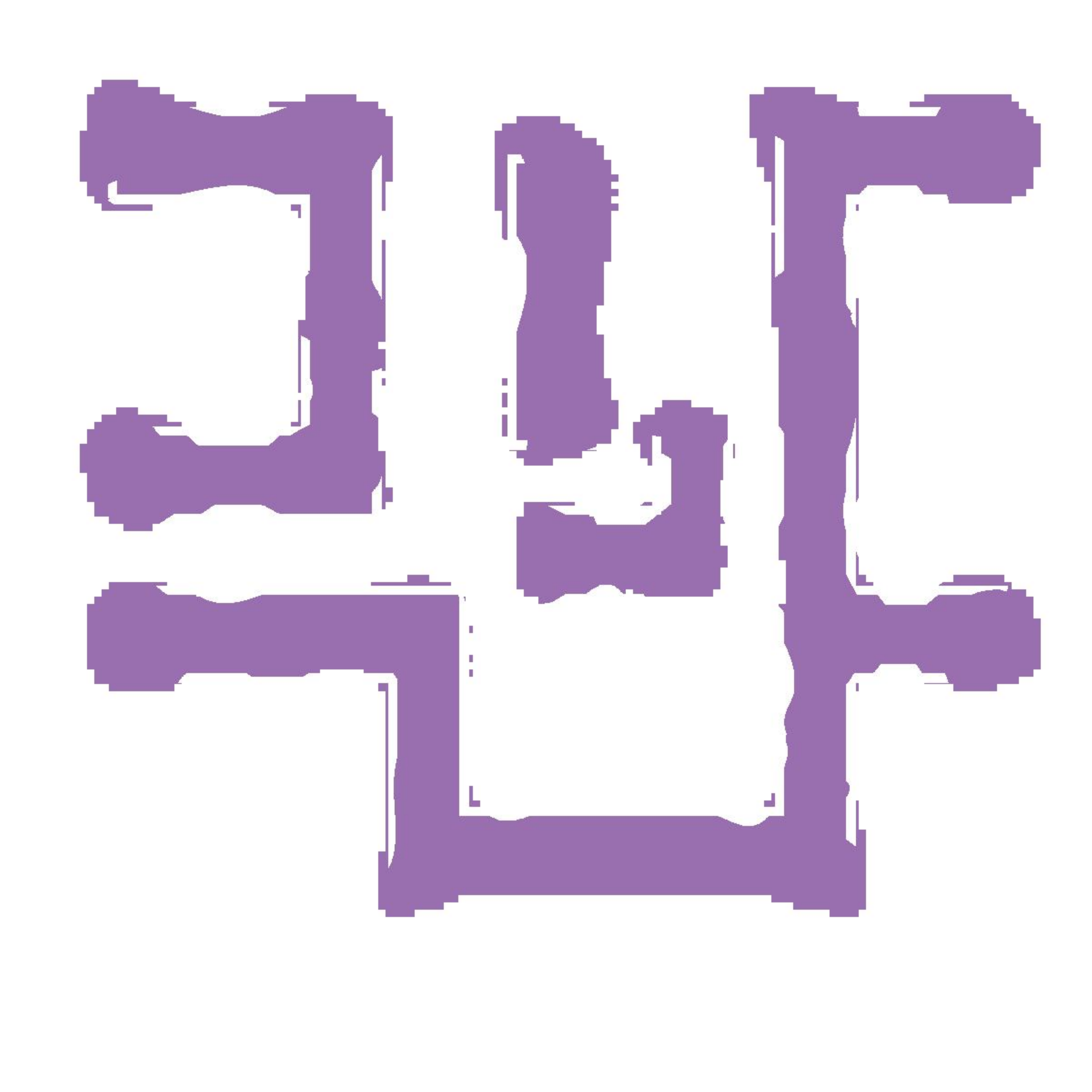} & \includegraphics[width=.21\linewidth,valign=m]{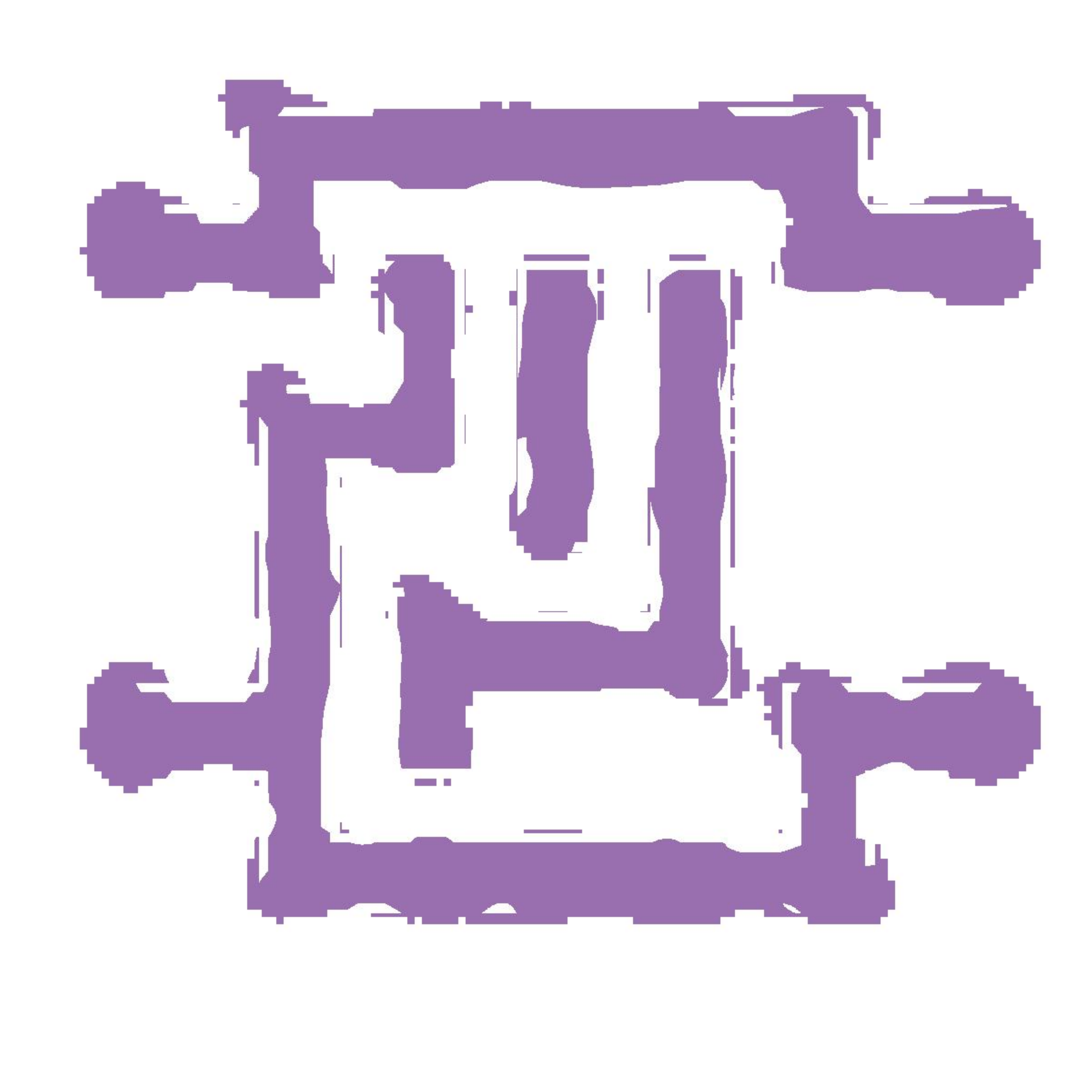} \\
    (b) GLS~\cite{OPC-TR2020-Yu}              & \includegraphics[width=.21\linewidth,valign=m]{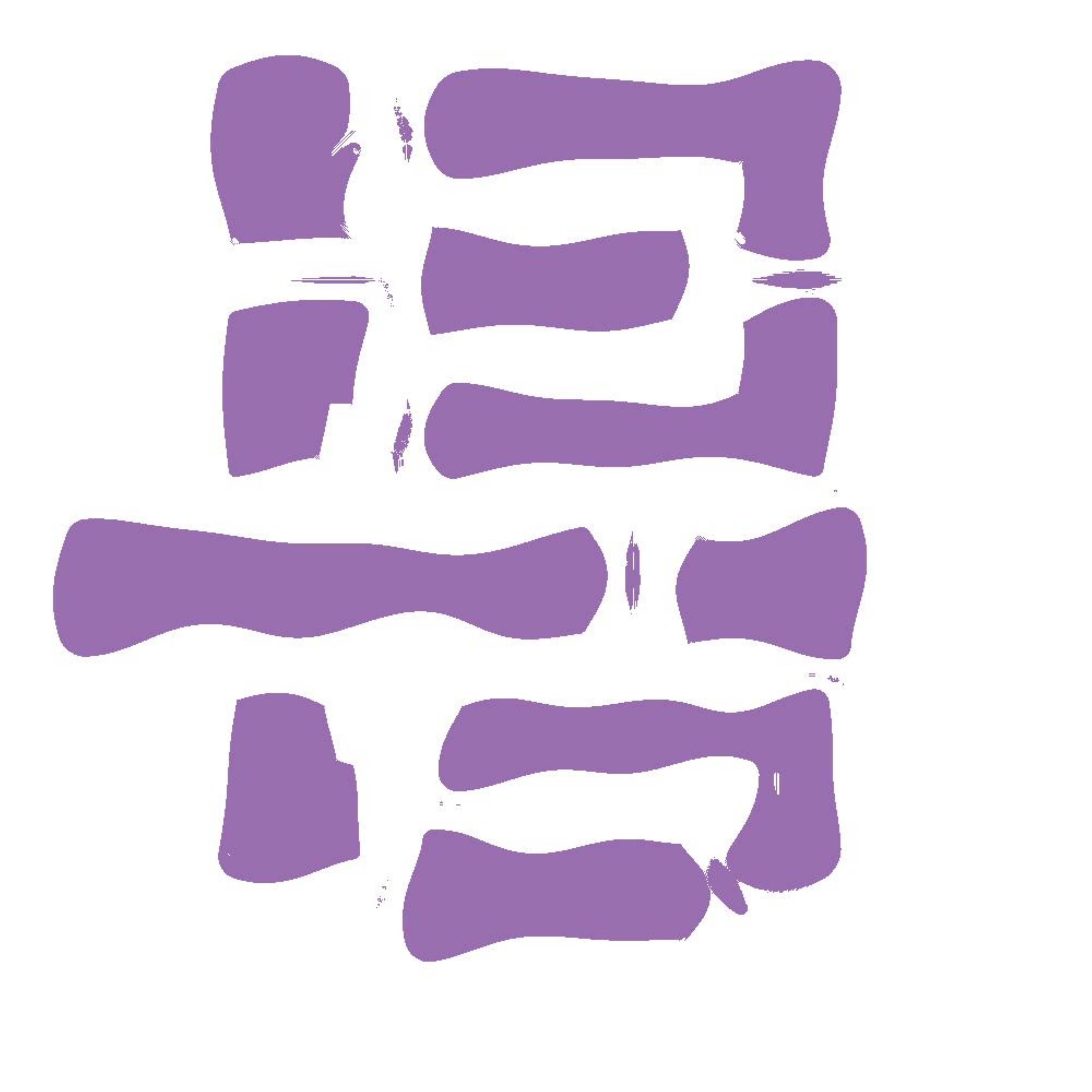}   & \includegraphics[width=.21\linewidth,valign=m]{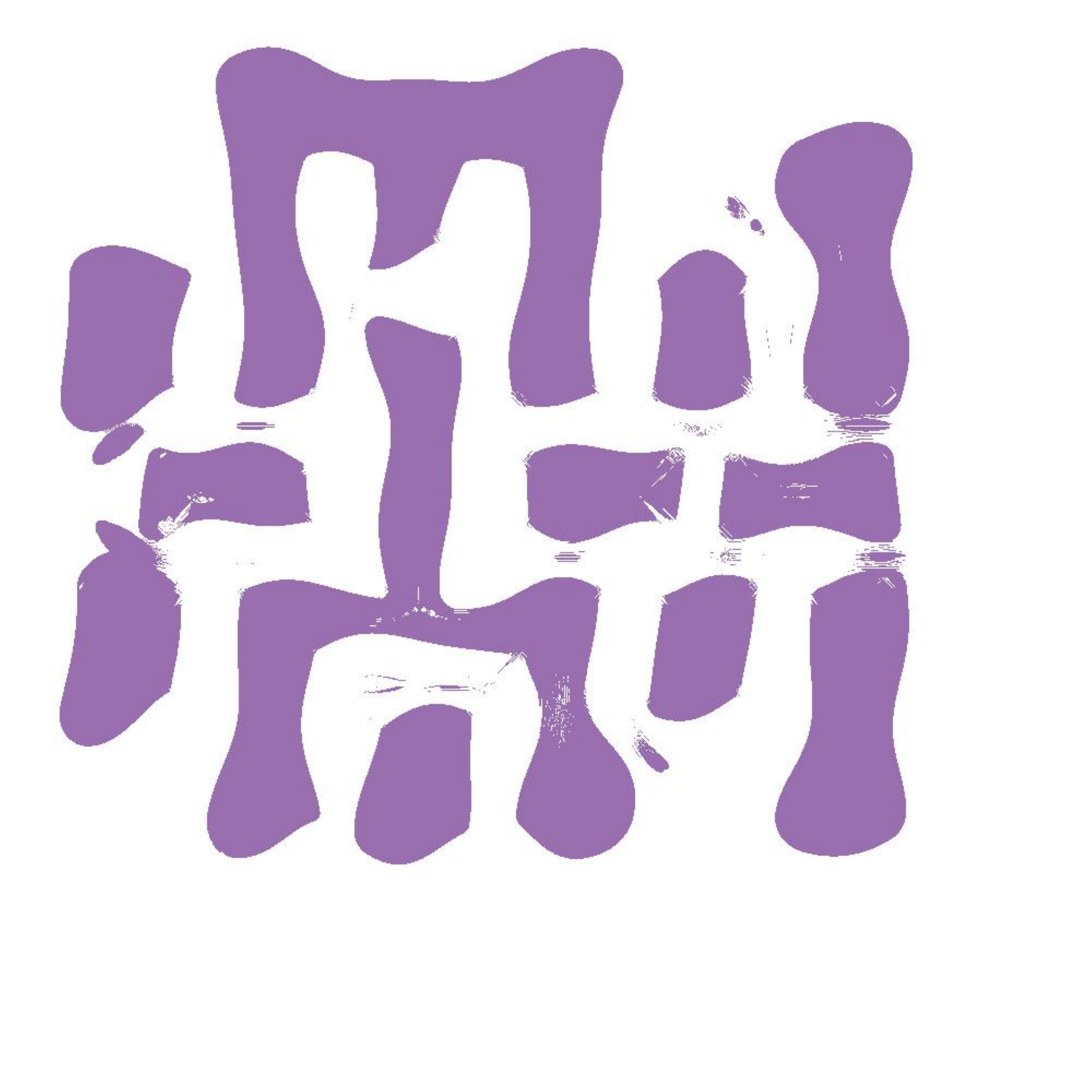}   & \includegraphics[width=.21\linewidth,valign=m]{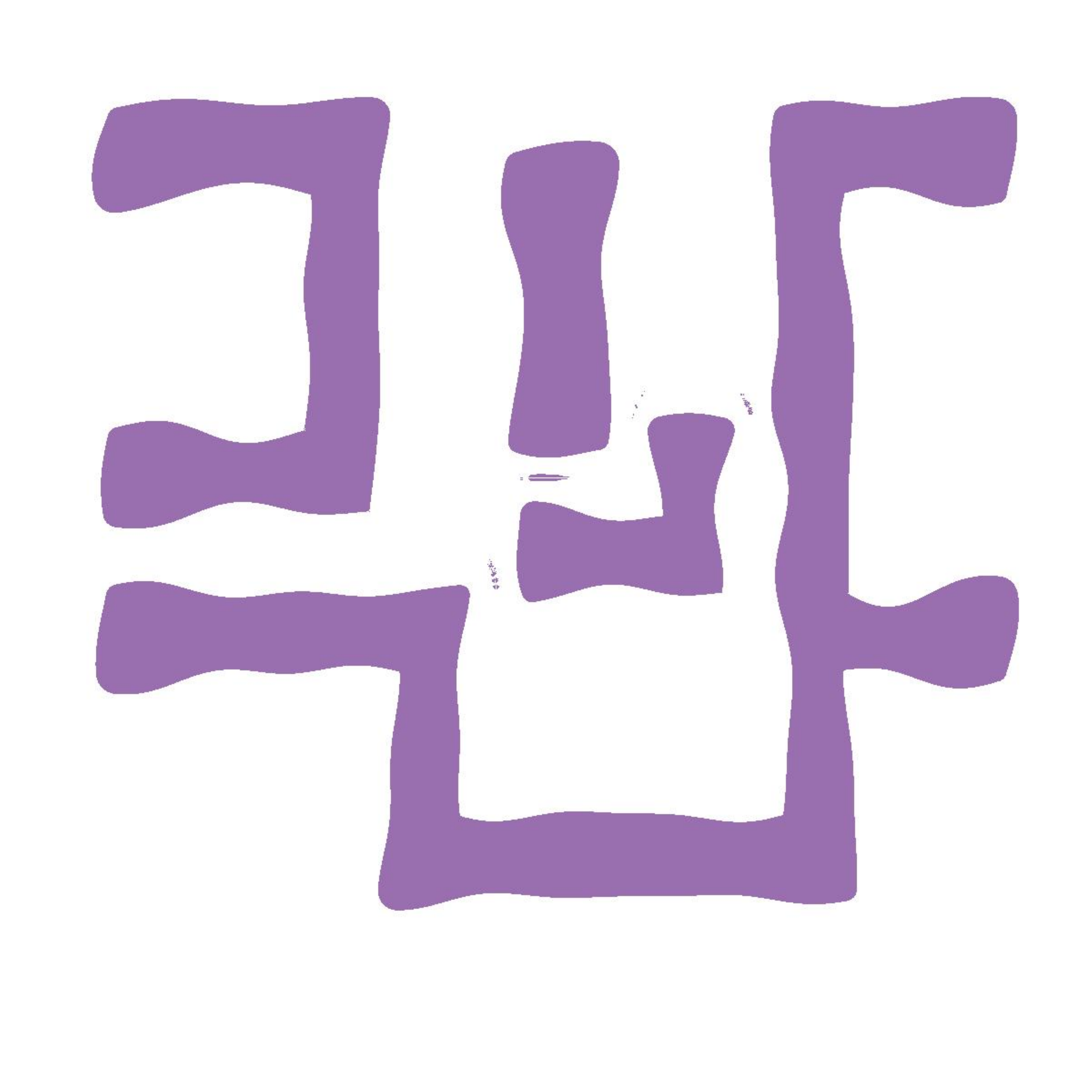}   & \includegraphics[width=.21\linewidth,valign=m]{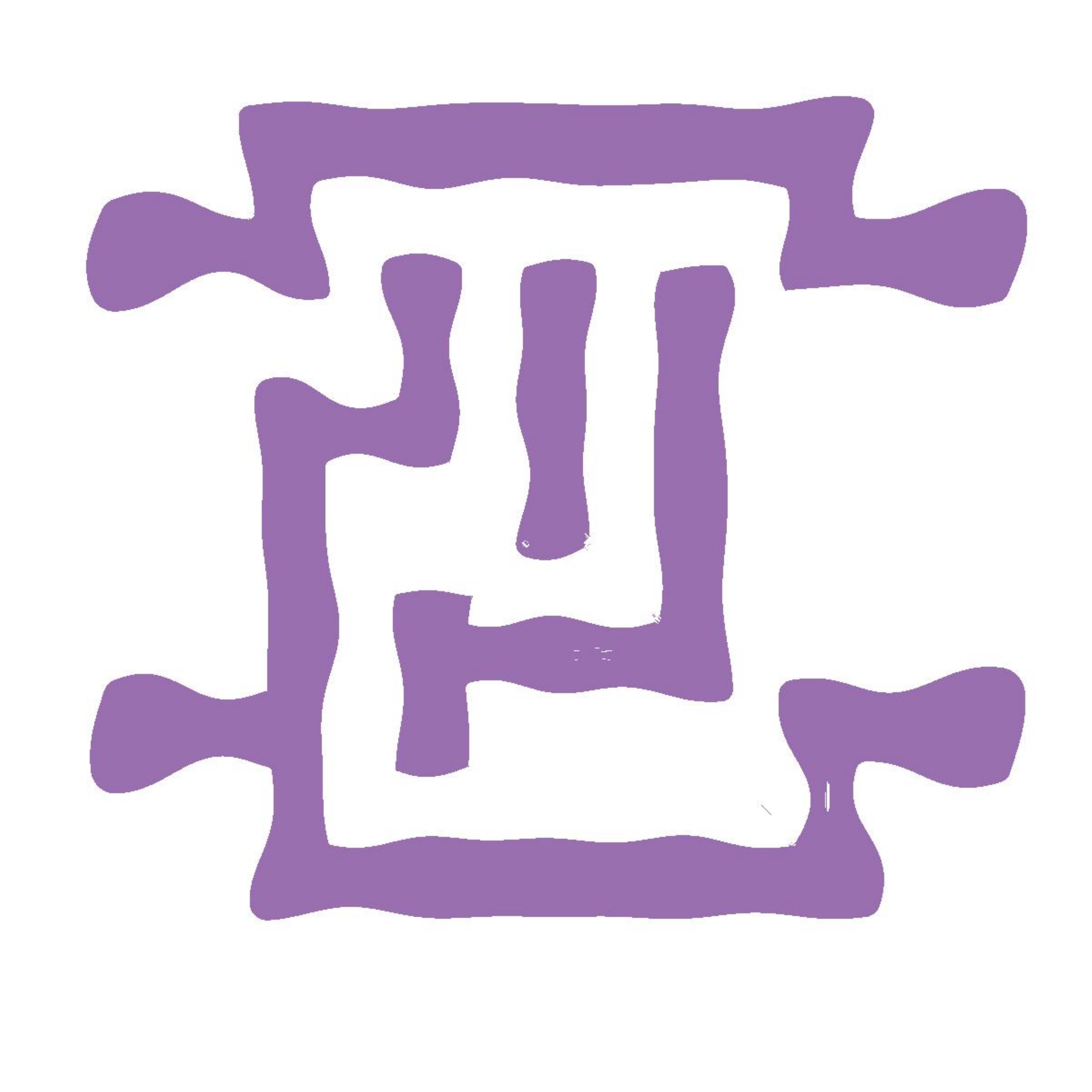}   \\
    (c) NILT~\cite{NEURAL-ILT-ICCAD2020-Jiang}& \includegraphics[width=.21\linewidth,valign=m]{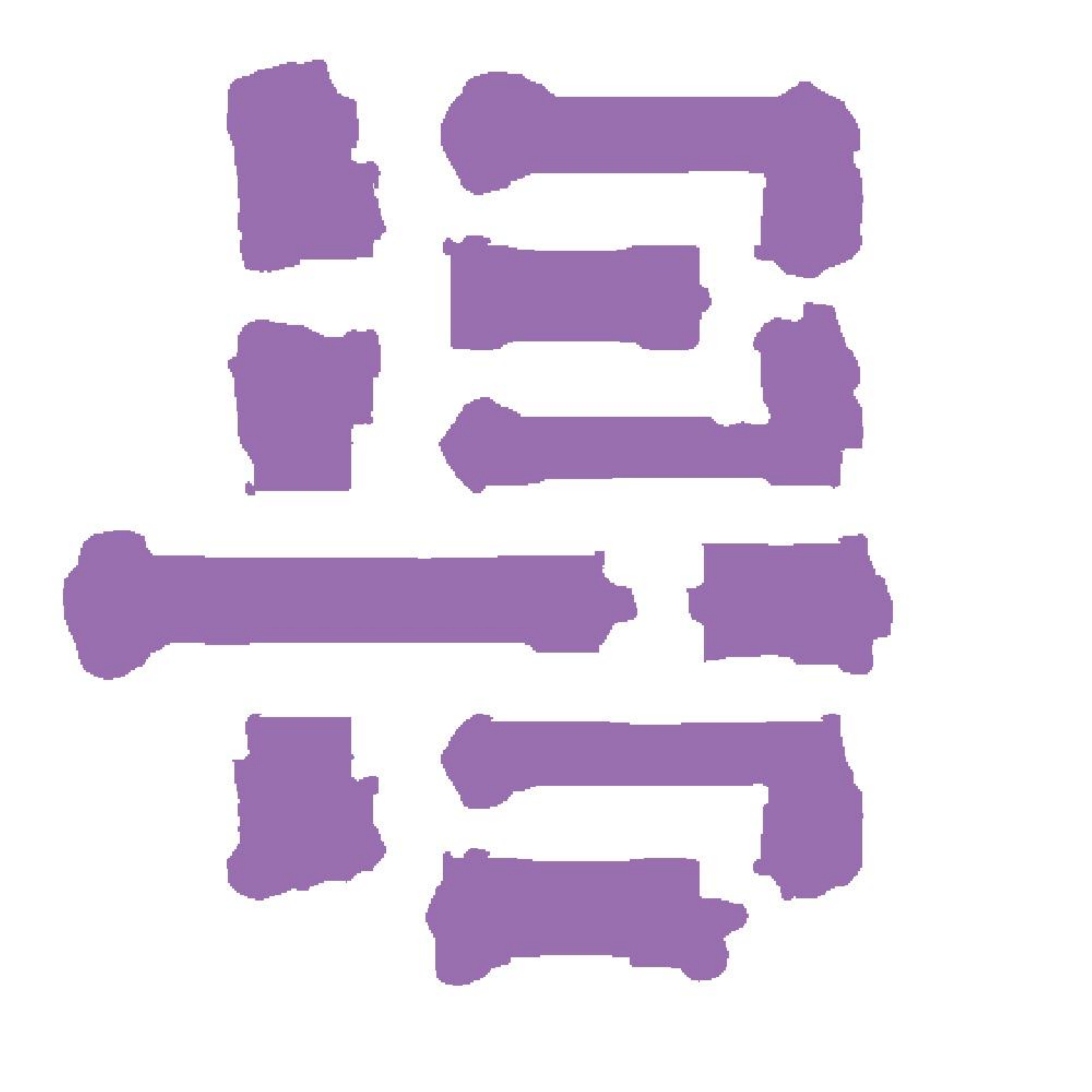}   & \includegraphics[width=.21\linewidth,valign=m]{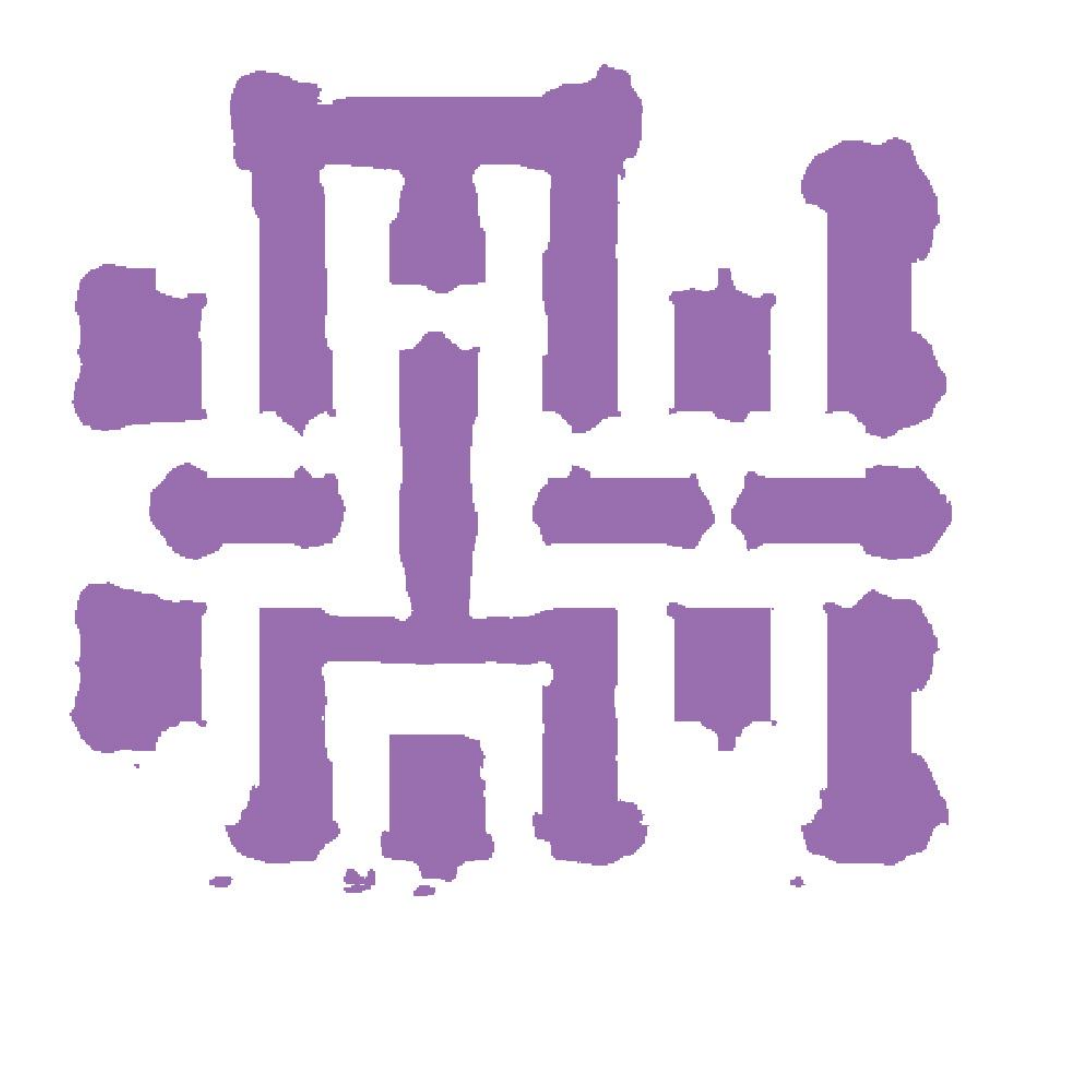}   & \includegraphics[width=.21\linewidth,valign=m]{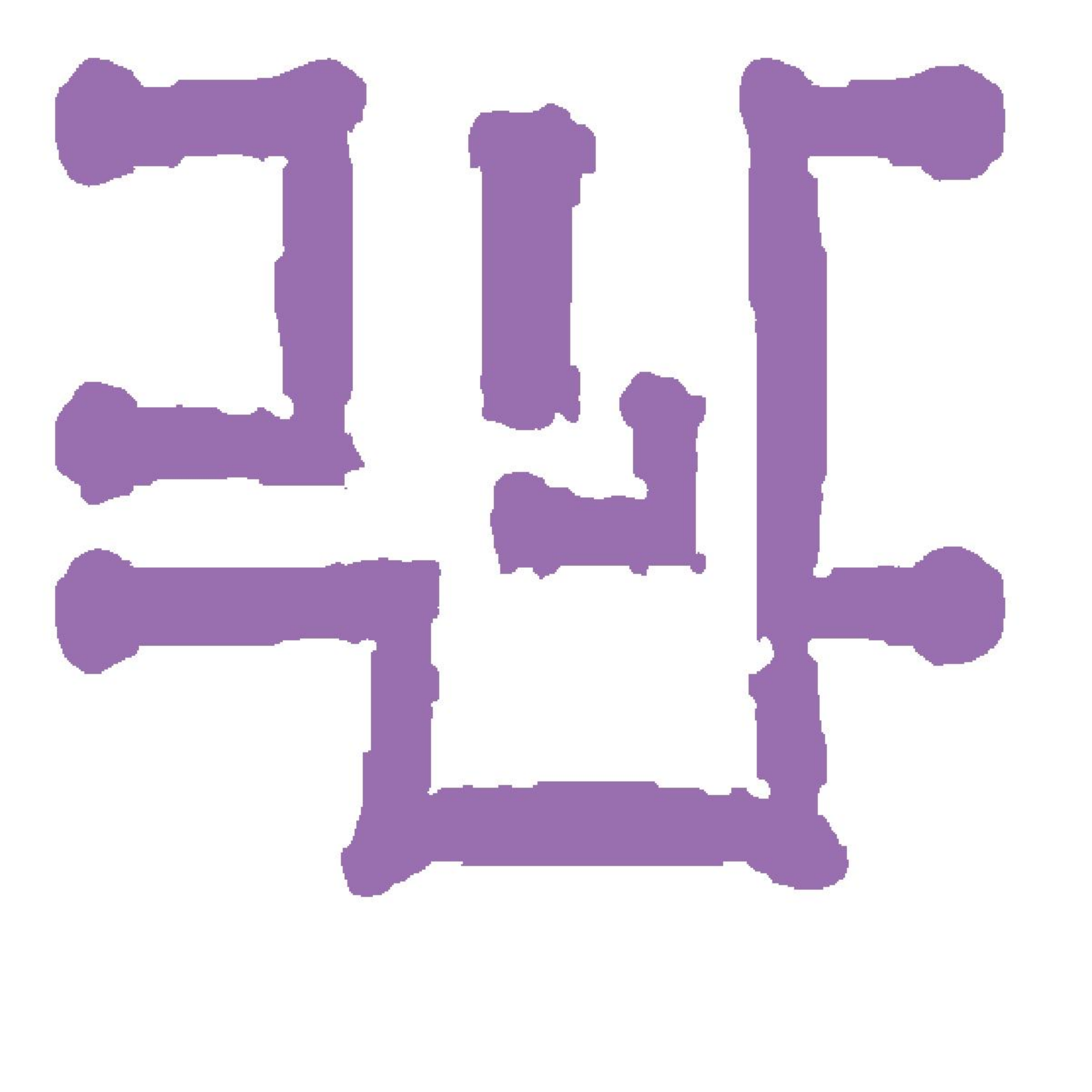}   & \includegraphics[width=.21\linewidth,valign=m]{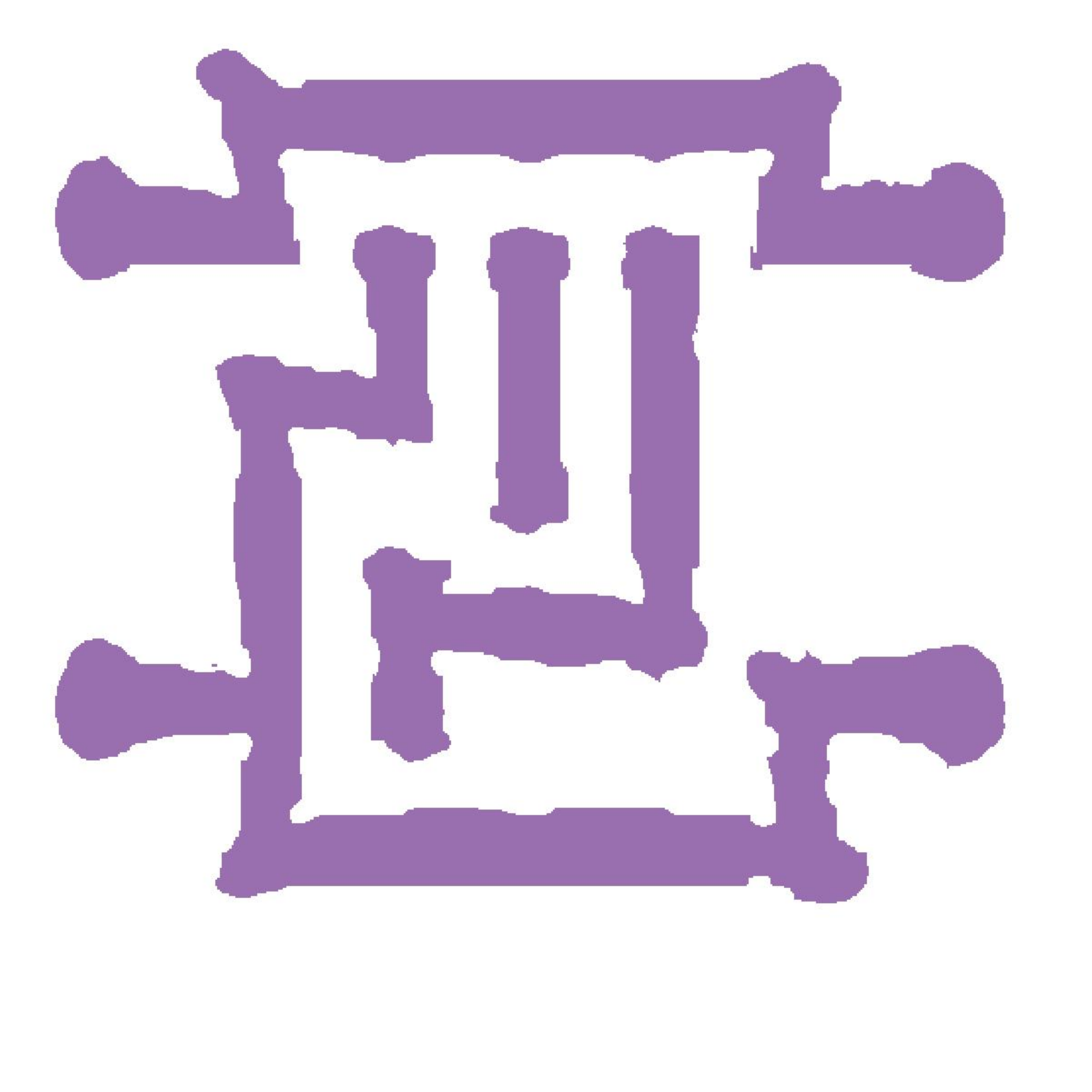} \\
    (d) DSO                                   & \includegraphics[width=.21\linewidth,valign=m]{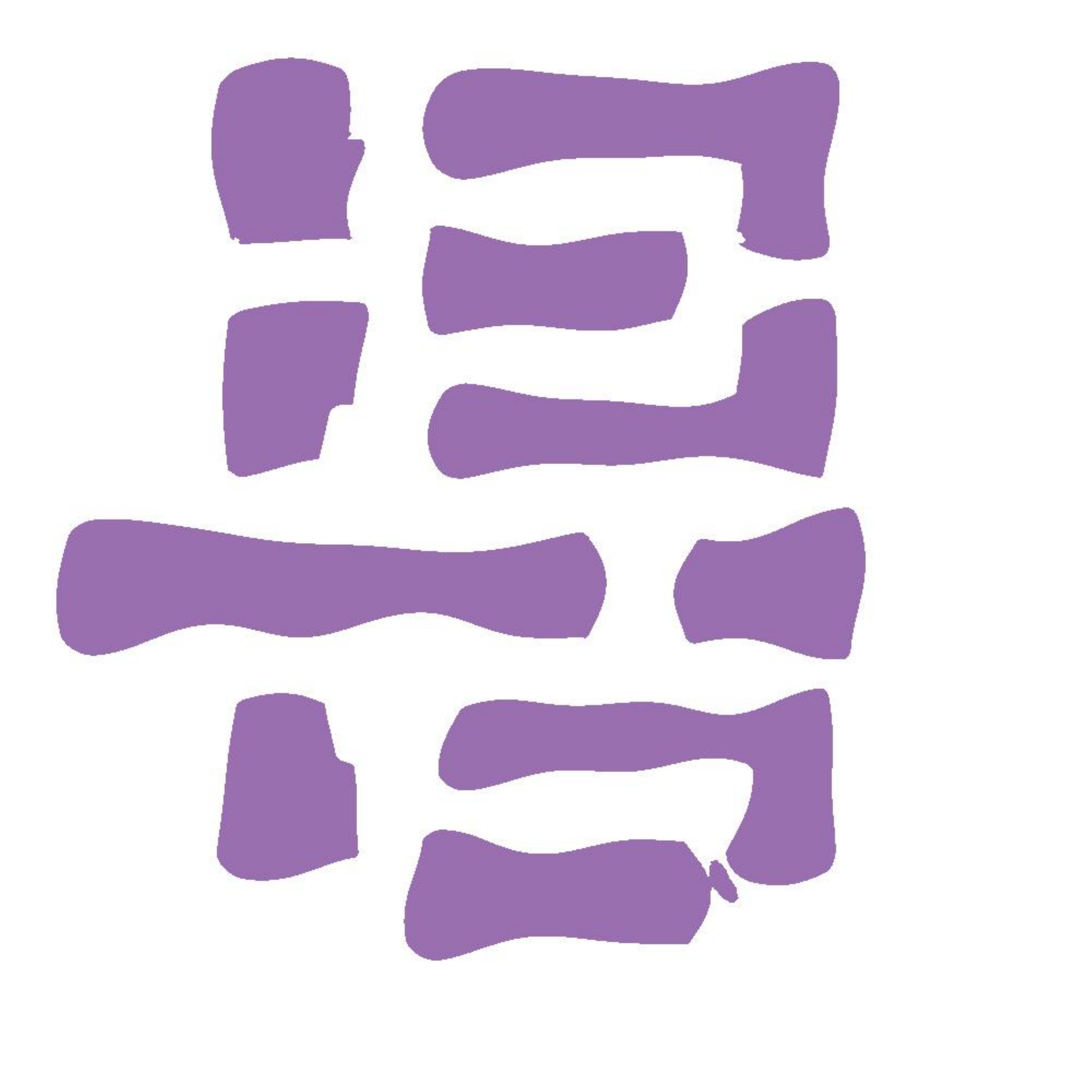}    & \includegraphics[width=.21\linewidth,valign=m]{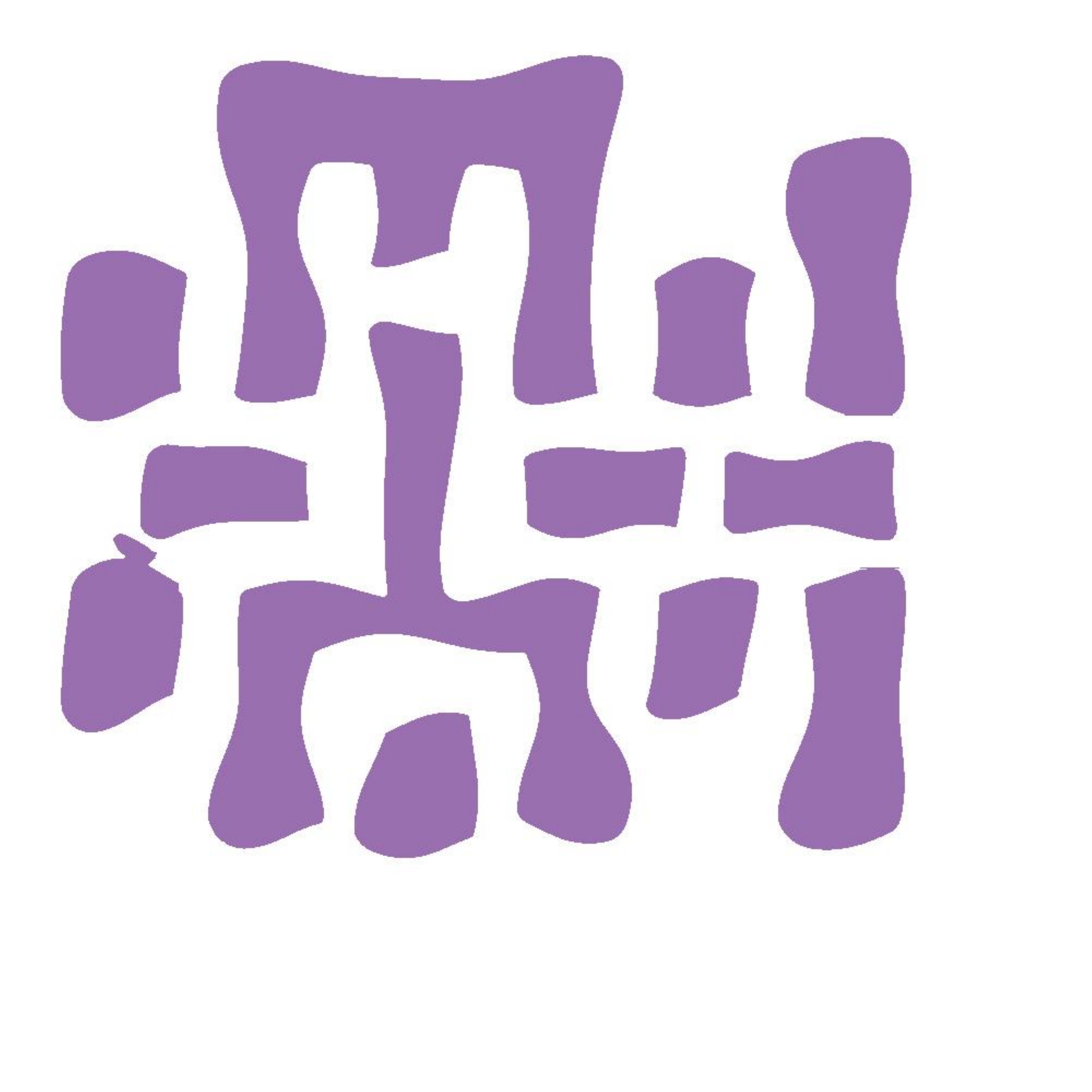}    & \includegraphics[width=.21\linewidth,valign=m]{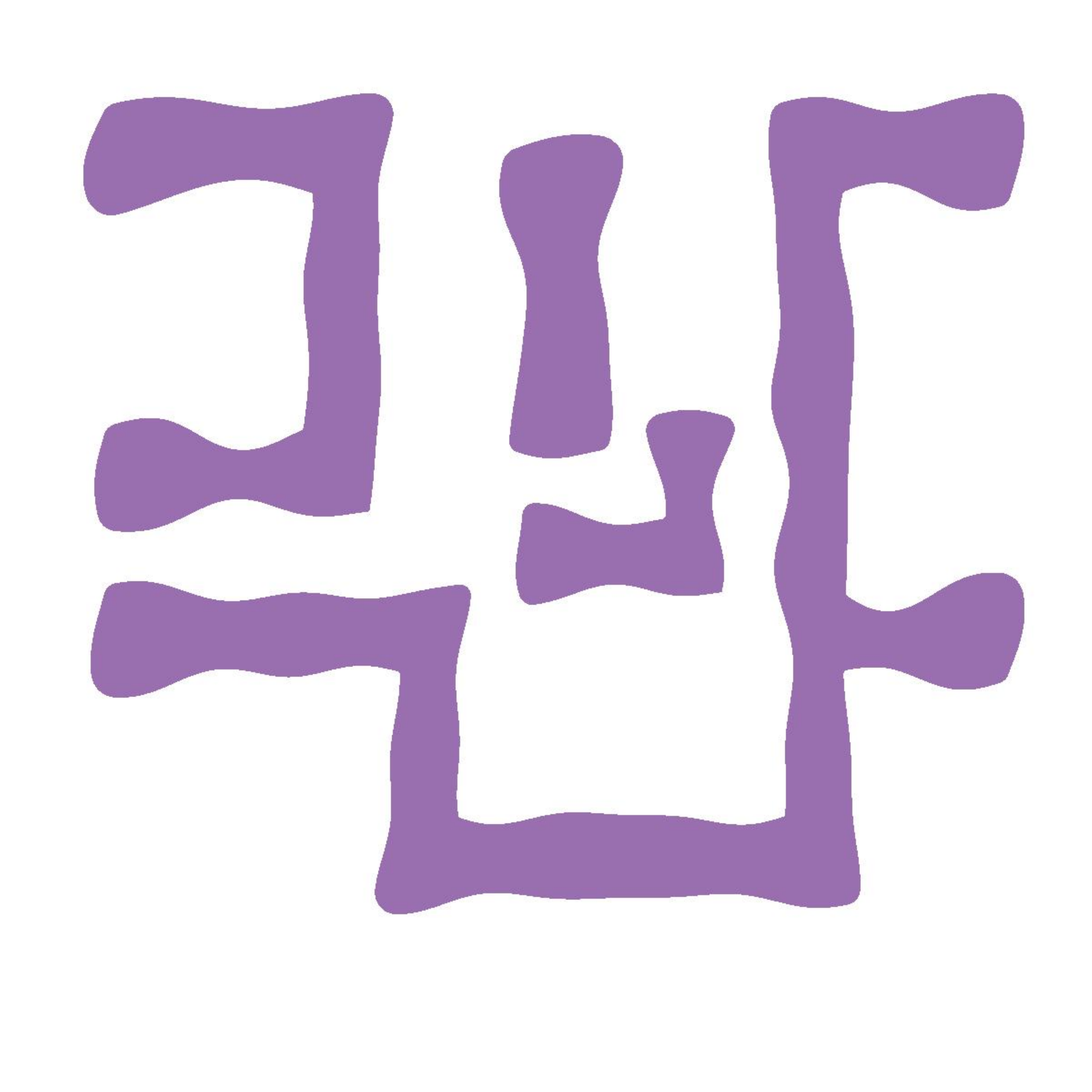}    & \includegraphics[width=.21\linewidth,valign=m]{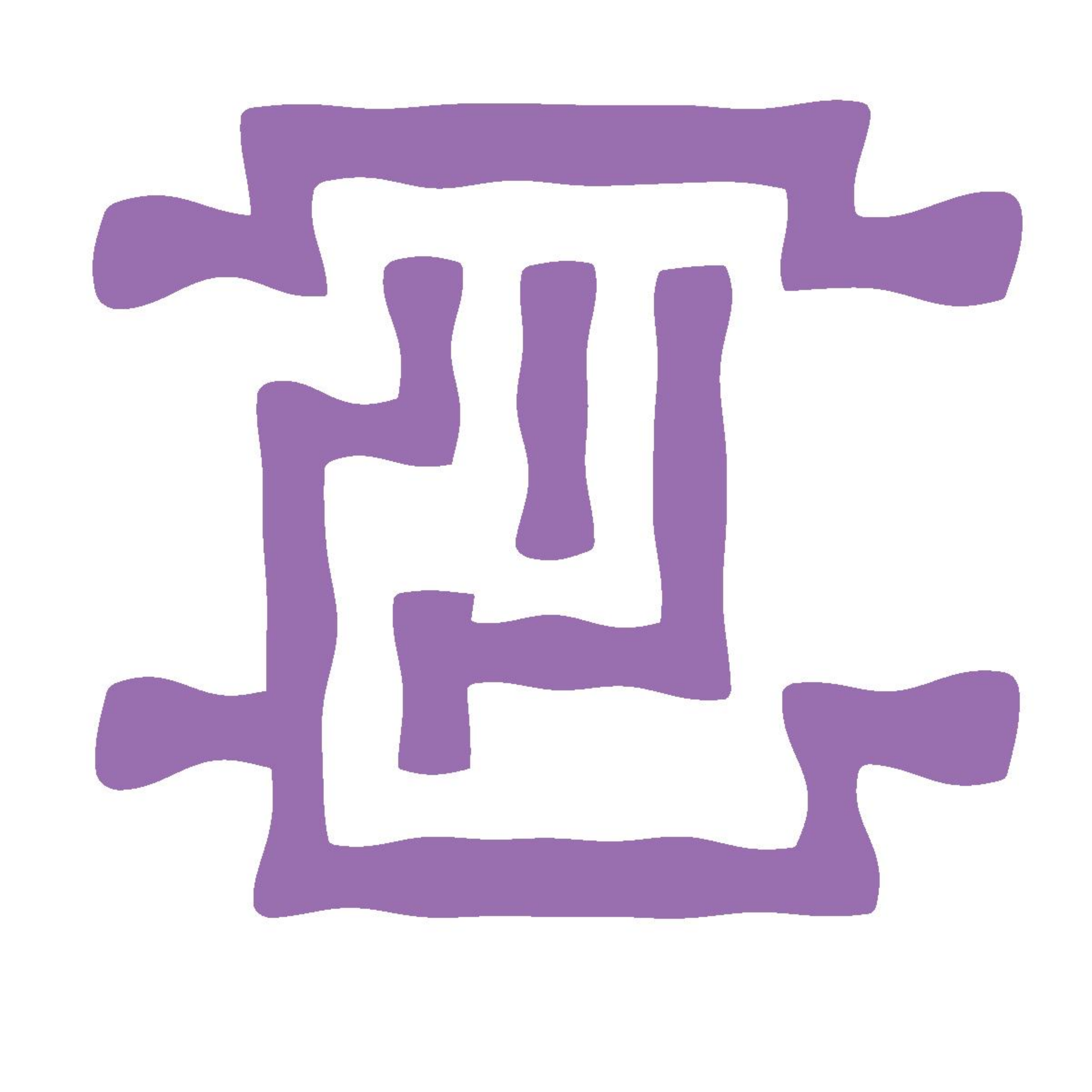}   \\
    (e) DevelSet                              & \includegraphics[width=.21\linewidth,valign=m]{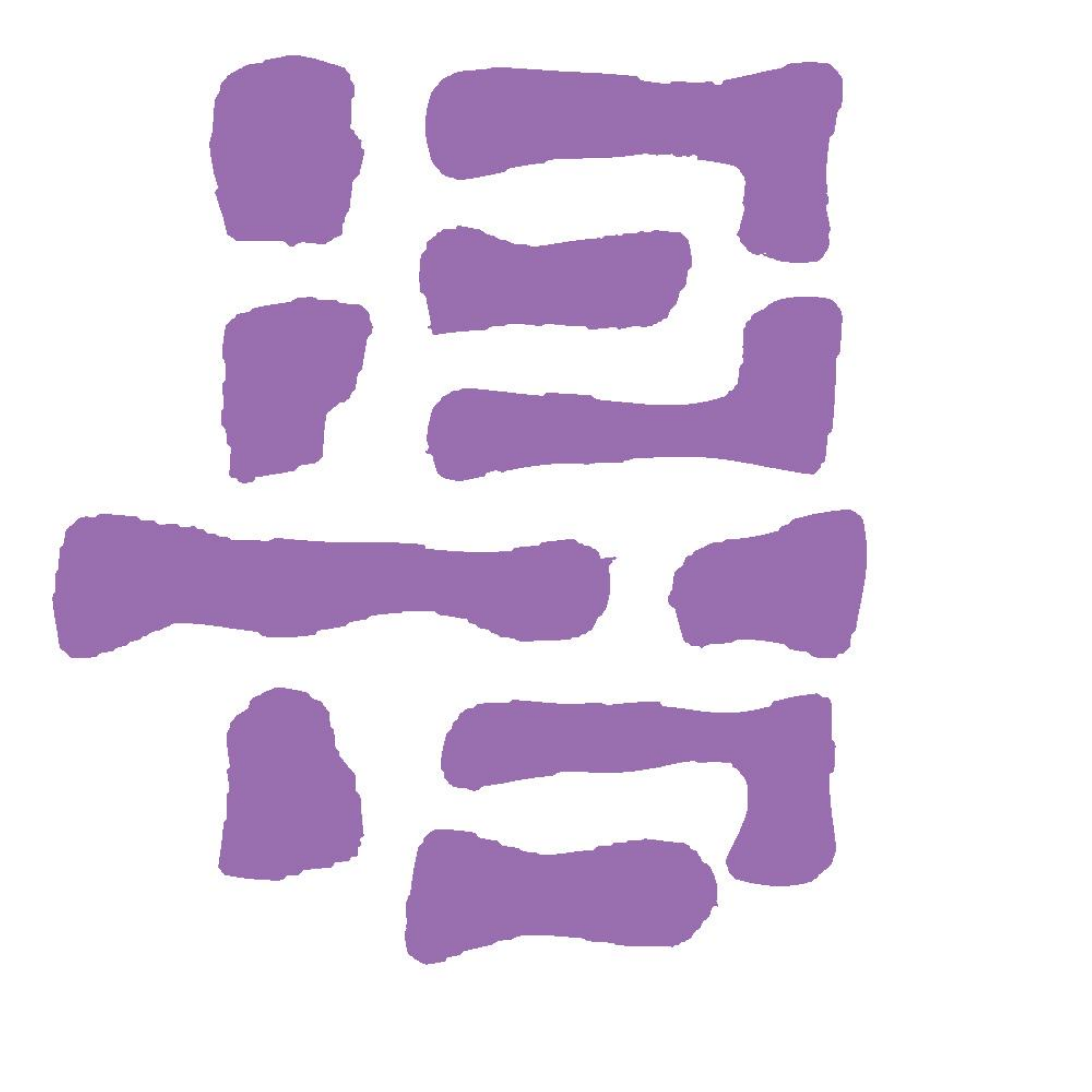}  & \includegraphics[width=.21\linewidth,valign=m]{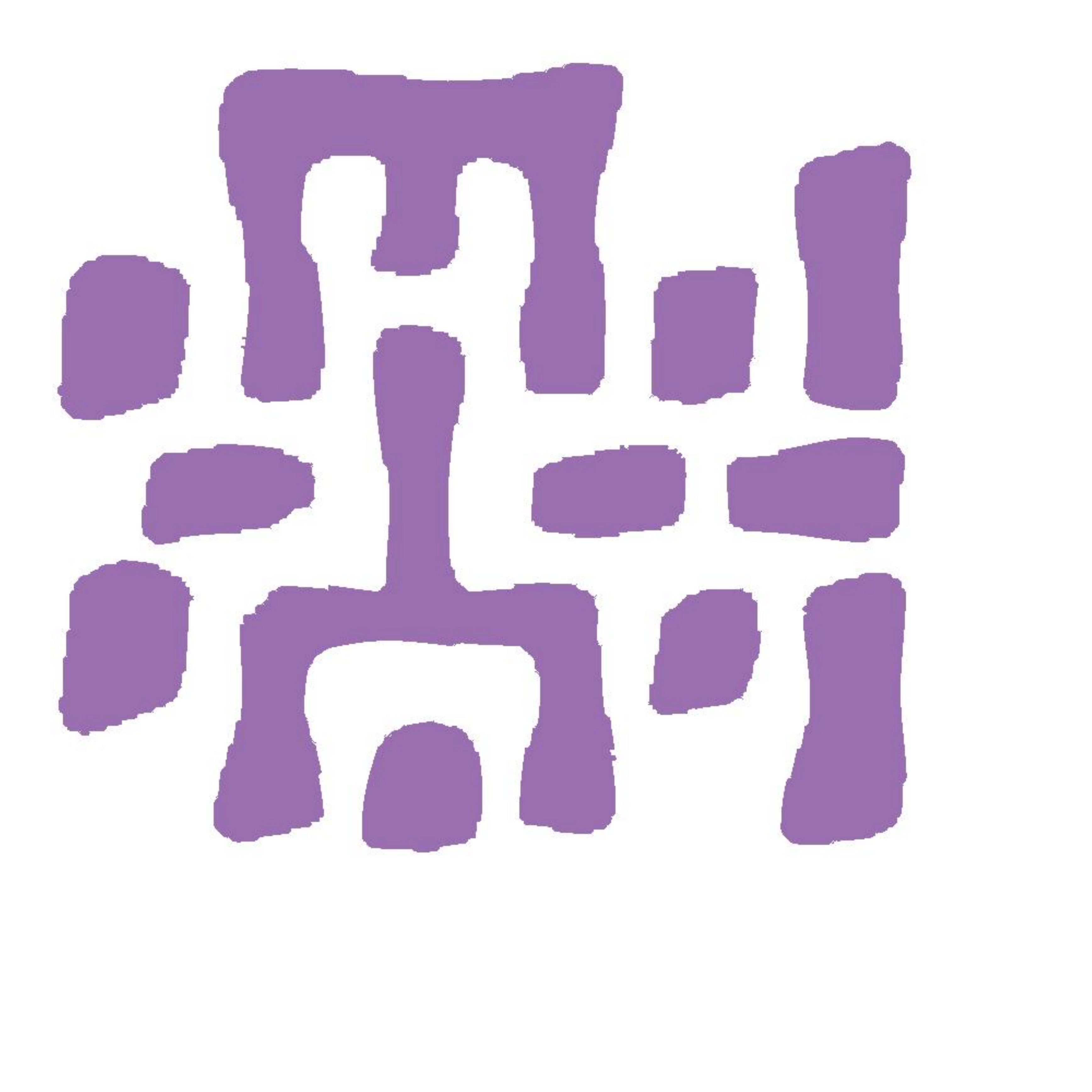}  & \includegraphics[width=.21\linewidth,valign=m]{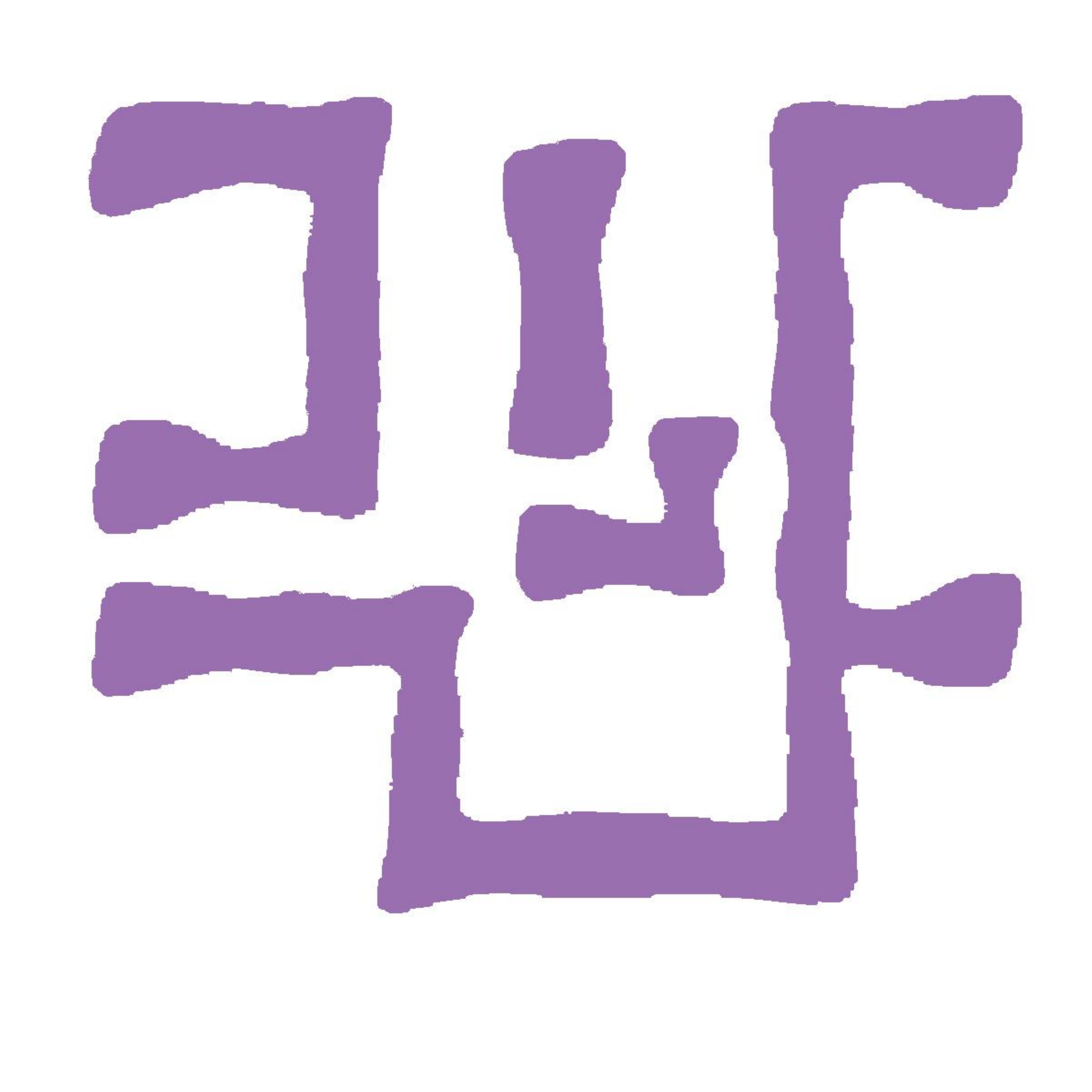}  & \includegraphics[width=.21\linewidth,valign=m]{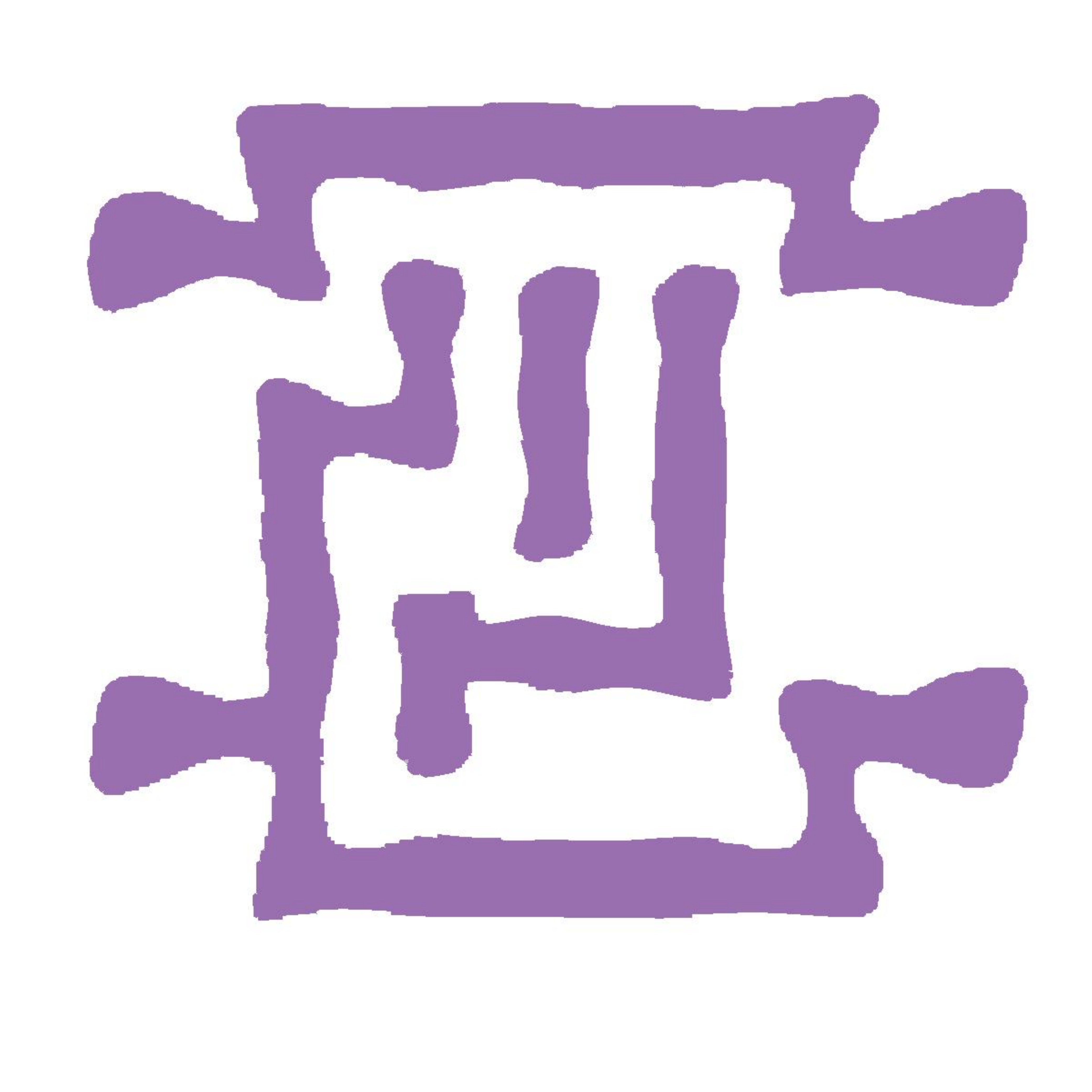} \\
  \end{tabular}
  \captionof{figure}{Mask visualizations of: (a) PGAN-OPC~\cite{OPC-TCAD2020-Yang},
  (b) GLS-ILT~\cite{OPC-TR2020-Yu}, (c) Neural-ILT~\cite{NEURAL-ILT-ICCAD2020-Jiang},
  (d) DSO, and (e) DevelSet framework (DSN + DSO).}
  \label{fig:vis_masks}
\end{table}

We compare the performance of the proposed DevelSet with other SOTA mask optimization methods,
and the detailed values are listed in \Cref{tab:develset_results}.
Compared with the conventional ILT, the $L_2$ and PVB are reduced by $14.6\%$ and $4.6\%$ respectively.
Compared with the conventional level set mask optimization GLS-ILT, the $L_2$ and PVB could reduce by $0.6\%$ and $2.8\%$.
Our framework also displays superiority when compared with PGAN-OPC and Neural-ILT, which are two high-performance machine learning-based mask optimization frameworks.
The performance of $L_2$ could achieve $4.0\%$ and $0.3\%$ improvements and PVB could obtain $2.6\%$ and $9.6\%$ improvements, respectively.
As shown in \Cref{fig:vis_masks}, these results prove the high quality of DevelSet generated masks.

Among the above-mentioned methods, the shot numbers of DevelSet reduced by $85.8\%$, $38.4\%$ and $1.0\%$ compared with ILT, GLS-OPC and PGAN-OPC.
For Neural-ILT which also considers mask complexity, DevelSet generated masks contain $20.2\%$ more shots.
As depicted in \Cref{fig:vis_masks}, although the masks of DSO and DevelSet contain fewer stains, more shots are needed to keep the boundaries smooth.
And the quality and simplicity of masks are a trade-off, we are more concerned about the mask printability, this performance is acceptable.

\begin{table}[tbp]
  \centering
  \caption{Runtime comparison with SOTA.}
  \label{tab:develset_tat_results}
  \setlength{\tabcolsep}{2.8pt}
  \renewcommand{\arraystretch}{1.2}
  \begin{tabular}{c|c|c|c|c|c|c}
      \toprule
      \multirow{2}{*}{Bench} & \multicolumn{1}{c|}{ ILT \cite{OPC-DAC2014-Gao} } & \multicolumn{1}{c|}{ GLS \cite{OPC-TR2020-Yu} } & \multicolumn{1}{c|}{ PGAN \cite{OPC-TCAD2020-Yang} } & \multicolumn{1}{c|}{ NILT \cite{NEURAL-ILT-ICCAD2020-Jiang} } & \multicolumn{1}{c|}{ DSO } & \multicolumn{1}{c}{ DevelSet }  \\
      & TAT (s) & TAT (s) & TAT (s) & TAT (s) & TAT (s) & TAT (s) \\ \midrule
      \texttt{case1} &1280 &123 &358 &13.57 &3.39 &\textbf{ 1.5 }    \\
      \texttt{case2} &381 &81 &368 &14.37 &2.84 &\textbf{ 1.4 }    \\
      \texttt{case3} &1123 &214 &368 &9.72 &3.59 &\textbf{ 1.29 }    \\
      \texttt{case4} &1271 &184 &377 &10.4 &4.1 &\textbf{ 1.65 }    \\
      \texttt{case5} &1120 &76 &369 &10.04 &2.68 &\textbf{ 0.91 }    \\
      \texttt{case6} &391 &65 &364 &11.11 &2.57 &\textbf{ 0.84 }    \\
      \texttt{case7} &406 &64 &377 &9.67 &2.32 &\textbf{ 0.76 }    \\
      \texttt{case8} &388 &67 &383 &11.81 &2.67 &\textbf{ 1.14 }    \\
      \texttt{case9} &1138 &63 &383 &9.68 &2.86 &\textbf{ 1.21 }    \\
      \texttt{case10} &387 &64 &366 &11.46 &2.27 &\textbf{ 0.42 }    \\   \midrule
      \multicolumn{1}{c|}{Average} &788.5 &100.1 &371.3 &11.18 &2.93 &\textbf{1.11} \\
      \multicolumn{1}{c|}{Ratio} &710.360 &90.180 &334.505 &10.072 &2.640 &\textbf{1.000} \\  \bottomrule
  \end{tabular}
\end{table}

\subsubsection{Runtime Comparison}
To prove the efficiency of our DevelSet mask optimization framework quantitively, we evaluate the turn around time (TAT) of different methods, as is shown in \Cref{tab:develset_tat_results}.
Compared with above-mentioned four methods, DevelSet could achieve significant speedup from $10\times$ to $710\times$.
With DSN as pre-processing, DevelSet can achieve $2.64\times$ speedup compared with DSO only,
this strongly proves the runtime performance improvement brought by DSN.

\subsection{Ablation Study.}
We conduct a set of ablation studies to evaluate the influence of each module in DevelSet.
As shown in \Cref{tab:abla_study}, we list the results of DSO with curvature term and without curvature term,
and the influence of modulation branch for end-to-end joint optimization of DevelSet (the DSN+DSO column).
The $L_2$, PVB, \#shots represent the square $L_2$ error, the area of PVBand, and the number of shots respectively.
We use the score to evaluate mask printability and mask complexity comprehensively, where the score is calculated as
the sum of the $L_2$, PVB, and $10 \times$\#shots.
The result is better if the score is smaller.

\begin{table}[tbp]
  \centering
  \caption{Ablation study.}
  \label{tab:abla_study}
  \setlength{\tabcolsep}{10pt}
  \renewcommand{\arraystretch}{1.2}
  \begin{tabular}{c|cc|cc}
  \toprule
            & \multicolumn{2}{c|}{DSO} & \multicolumn{2}{c}{DSN+DSO} \\
            & w/o.~curv.  & w.~curv.    & w/o.~mod.      & w.~mod.     \\ \midrule
            $L_2$      &  38253.0  &  38454.0  &  39259.8  &  38402.8 \\
            PVB        &  49243.0  &  49398.0  &  48384.0  &  48673.0 \\
            \#shots    &  805.0  &  726.0  &  712.8  &  699.8 \\ \midrule
            score$^\dagger$     &  95546.0  &  95112.0  &  94771.8  &  \textbf{94073.8} \\ \bottomrule
            \multicolumn{4}{l}{\textit{\ $^{\dagger}$\footnotesize{score = $L_2$ + PVB + 10 $\times$ \#shots}.}}
  \end{tabular}
\end{table}

\ifshowfig
\begin{table}
  \centering
  \setlength{\tabcolsep}{2pt}
  \begin{tabular}{c|c|c|c|c|c|c|c}
    \toprule
    w/o.~curv. & \includegraphics[width=.1\linewidth,valign=m]{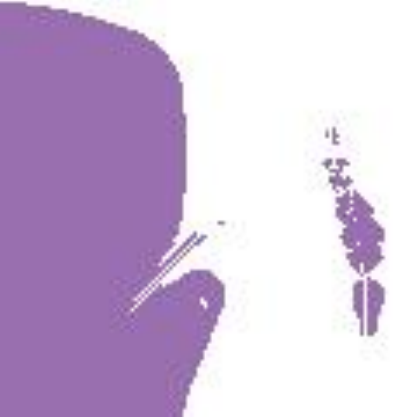}   & \includegraphics[width=.1\linewidth,valign=m]{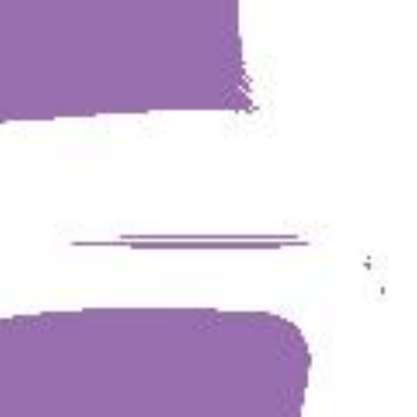} & \includegraphics[width=.1\linewidth,valign=m]{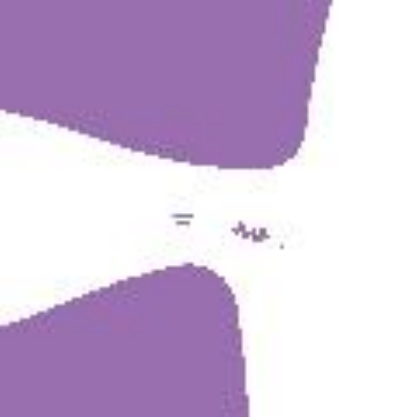} & \includegraphics[width=.1\linewidth,valign=m]{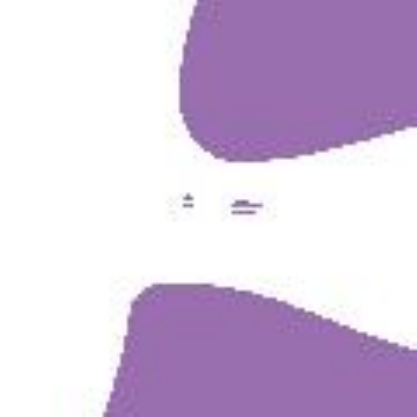} & \includegraphics[width=.1\linewidth,valign=m]{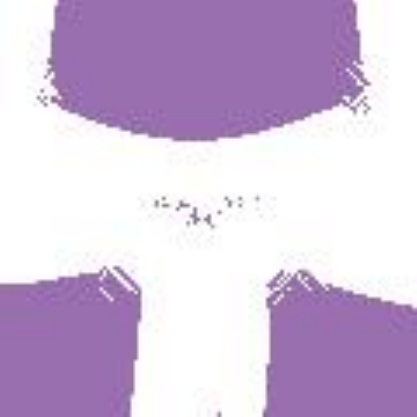} & \includegraphics[width=.1\linewidth,valign=m]{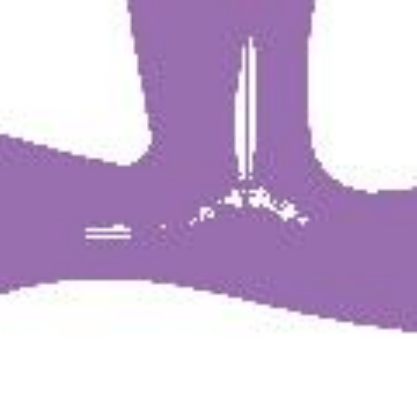} & \includegraphics[width=.1\linewidth,valign=m]{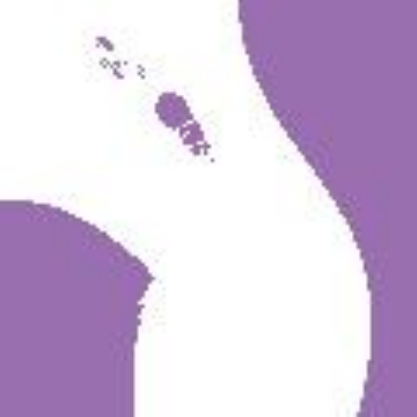} \\ \midrule
    w.~curv.  & \includegraphics[width=.1\linewidth,valign=m]{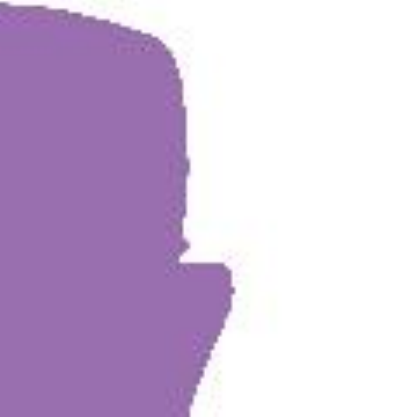}    & \includegraphics[width=.1\linewidth,valign=m]{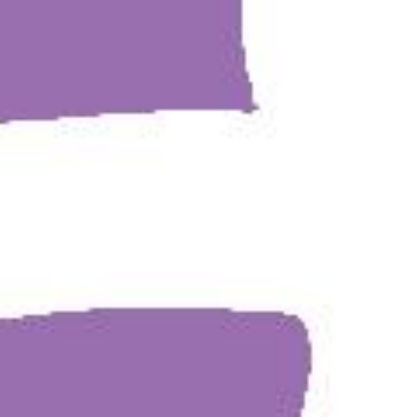}  & \includegraphics[width=.1\linewidth,valign=m]{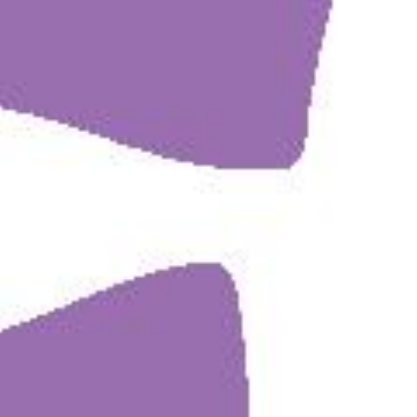}  & \includegraphics[width=.1\linewidth,valign=m]{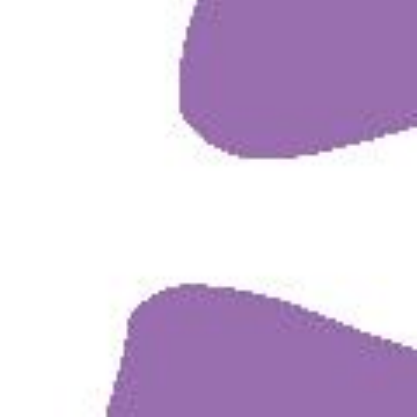}  & \includegraphics[width=.1\linewidth,valign=m]{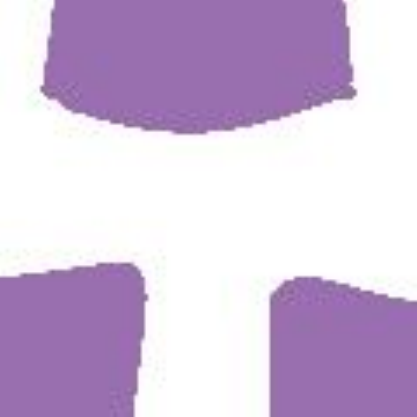}  & \includegraphics[width=.1\linewidth,valign=m]{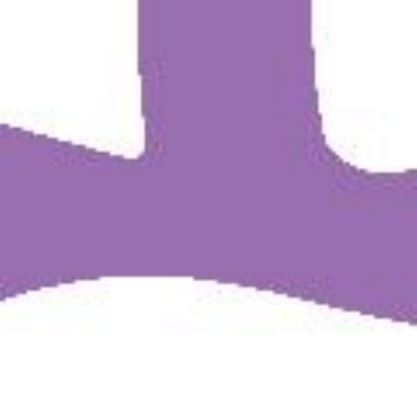}  & \includegraphics[width=.1\linewidth,valign=m]{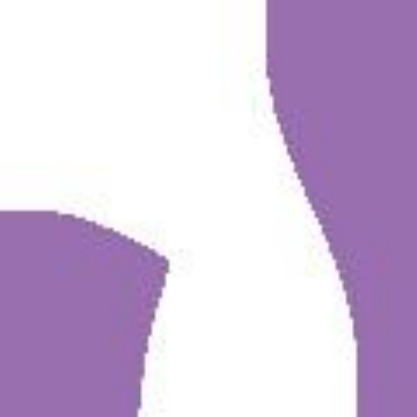}   \\ \bottomrule
  \end{tabular}
  \captionof{figure}{Visualizations for ablation study of the curvature term.}
  \label{fig:vis_curv}
\end{table}
\fi

\subsubsection{The Effectiveness of Curvature Term}
We analyze the influence of the curvature term on mask complexity and printability.
As the data listed in the DSO column in \Cref{tab:abla_study},
The DSO gets 805 \#shots without curvature term but reduces to 726 when with the guidance of the curvature.
In \Cref{fig:vis_curv}, we compare several parts of different masks to illustrate the influence of the curvature term.
As shown in figures, the curvature term makes the mask boundaries more smooth and eliminates the isolated stains and edge glitches.
Although the $L_2$ and PVB all become somewhat worse, which confirms our assumption that the curvature will harm mask printability a bit,
the total score drops indicating the loss caused by the curvature term is acceptable.

\subsubsection{The Necessity of DSN and Modulation Branch}
Comparing the column DSO and the DSN+DSO in \Cref{tab:abla_study},
we find that the DSN boosts the overall performance of DevelSet by the end-to-end joint optimization.
The DSN provides better initial LSF which help the DSO overcome the local minima and obtain better masks.
The \#shots number also drops because the upsample functions of the neural network make the mask generated by DSN more regular.

Further, we apply the modulation branch with curvature term to improve the overall score,
the result in DSN+DSO with mod. column reveals that the modulation branch improves mask printability and reduces complexity.
With the modulation branch, we have maximized the benefits of the curvature term while minimizing its adverse impacts.

\section{Conclusion}
\label{sec:conclu}
In this paper, we present DevelSet,
a CUDA and DNN accelerated end-to-end level set OPC framework that can perform instant mask optimization in around 1 second.
By introducing the curvature term into the level set algorithm, we extend the applicable scenarios of level set-based ILT methodology for mask manufacturability improvement.
Moreover, a novel multi-branch neural network with level set embeddings is proposed to boost the fast convergence of DevelSet.
We believe the improved level set algorithm with CUDA/DNN accelerated joint optimization paradigm
will have a real impact on the industrial mask optimization solutions.

\section*{Acknowledgment}

This work is partially supported by
Research Grants Council of Hong Kong SAR CUHK14209420, CUHK14208021.

\newpage
{
    \bibliographystyle{IEEEtran}
    \bibliography{./ref/Top,./ref/HSD,./ref/DFM,./ref/DL,./ref/PD,./ref/Software,./ref/LS}
}

\end{document}